\documentclass[10pt,twocolumn,letterpaper]{article}

\usepackage{wacv}
\usepackage{times}
\usepackage{epsfig}
\usepackage{graphicx}
\usepackage{amsmath}
\usepackage{amssymb}

\usepackage{comment}
\usepackage{color}
\usepackage{siunitx}
\usepackage{booktabs}
\usepackage{bm}
\usepackage{mathtools}
\usepackage{floatrow}
\usepackage{multirow}
\usepackage{wrapfig,lipsum,booktabs}
\usepackage{caption}
\usepackage{subcaption}
 \usepackage{sidecap}
\usepackage{xspace}
\usepackage{textcomp}
\usepackage{relsize}
\usepackage[usenames,dvipsnames]{xcolor}

\usepackage{enumitem}
\setlist{nosep}

\usepackage{xspace}
\usepackage[export]{adjustbox}

\usepackage{bold-extra}

\newcommand{\reffig}[1]{Fig.\,\ref{fig:#1}}
\newcommand{\reftab}[1]{Tab.\,\ref{tab:#1}}

\makeatletter

\definecolor{ubpubColor}{rgb}{0.43, 0.5, 0.5}
\definecolor{backboneColor}{rgb}{0.423, 0.325, 0.365}
\definecolor{fpnColor}{rgb}{0.255, 0.498, 0.416}

\newcommand{\PAR}[1]{\vskip3pt \noindent {\bf #1~}}

\newcommand{\vect}[1]{\mathbf{#1}}

\newcolumntype{P}[1]{>{\centering\arraybackslash}p{#1}}

\usepackage{pifont}
\newcommand{\cmark}{\ding{51}}%
\newcommand{\xmark}{\ding{55}}%

\newcommand{\J}{\mathcal{J}}
\newcommand{\F}{\mathcal{F}}
\newcommand{\JnF}{\mathcal{J}\&\mathcal{F}}

\definecolor{bistre}{rgb}{0.24, 0.17, 0.12}
\definecolor{bluepigment}{rgb}{0.2, 0.2, 0.6}

\definecolor{irrelColor}{rgb}{0.43, 0.5, 0.5}
\newcommand{\IRREL}[1]{\textcolor{irrelColor}{#1}}

\renewcommand{\textrightarrow}{$\rightarrow$}

\wacvfinalcopy 

\usepackage[breaklinks=true,bookmarks=false]{hyperref}

\begin{document}

\newcommand{\OurConvName}[0]{$\text{D}^2\text{Conv3D}$\xspace}

\title{$\text{D}^2 \text{Conv3D}$: Dynamic Dilated Convolutions for Object Segmentation in Videos}

\author{Christian Schmidt
\qquad
Ali Athar
\qquad
Sabarinath Mahadevan
\qquad
Bastian Leibe\\
Computer Vision Group, RWTH Aachen University\\
{\tt\small christian.schmidt4@rwth-aachen.de}\qquad
{\tt\small \{athar,mahadevan,leibe\}@vision.rwth-aachen.de}
}

\maketitle
\thispagestyle{empty}

\begin{abstract}
   Despite receiving significant attention from the research community, the task of segmenting and tracking objects in monocular videos still has much room for improvement. 
   Existing works have simultaneously justified the efficacy of dilated and deformable convolutions for various image-level segmentation tasks. This gives reason to believe that 3D extensions of such convolutions should also yield performance improvements for video-level segmentation tasks. 
   %
   However, this aspect has not yet been explored thoroughly in existing literature.
   In this paper, we propose Dynamic Dilated Convolutions (\OurConvName): a novel type of convolution which draws inspiration from dilated and deformable convolutions and extends them to the 3D (spatio-temporal) domain.
   We experimentally show that \OurConvName can be used to improve the performance of multiple 3D CNN architectures across multiple video segmentation related benchmarks by simply employing \OurConvName as a drop-in replacement for standard convolutions. 
   We further show that \OurConvName out-performs trivial extensions of existing dilated and deformable convolutions to 3D.
   Lastly, we set a new state-of-the-art on the DAVIS 2016 Unsupervised Video Object Segmentation benchmark.
   Code is made publicly available at \url{https://github.com/Schmiddo/d2conv3d}.
\end{abstract}

\section{Introduction}
The task of segmenting objects from monocular video sequences has received significant attention from the research community in recent years, mainly because of useful applications in self-driven cars, autonomous robots, \etc. Several existing approaches~\cite{Luiten18ACCV,Zulfikar19CVPRW,Bertasius20CVPR} for this task follow a two-step paradigm where objects are first segmented in individual image frames, followed by a second temporal association step.
Such methods leveraged the availability of accurate image-level instance segmentation networks \cite{Pinheiro16ECCV,He17ICCV} and various cues for temporal association (\eg attention, optical flow, Re-ID)~\cite{Luiten18ACCV,Zulfikar19CVPRW,Yang19ICCVAnchorDiff,Ren21CVPR}.
More recently however, methods have emerged~\cite{Hou19BMVC,Athar20ECCV,Mahadevan20BMVC} which use 3D convolutions to jointly reason over spatial and temporal dimensions, resulting in improved performance for various video object segmentation related tasks. 

\begin{figure}[t]
    \centering
    \includegraphics[width=\textwidth]{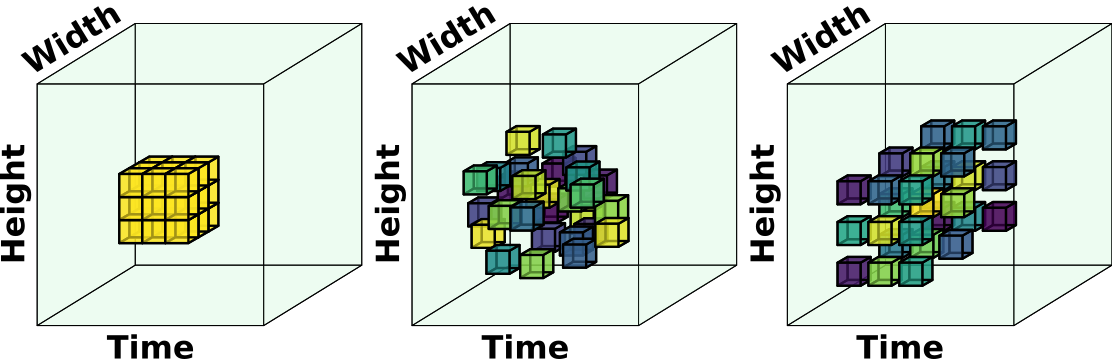}
    \caption{
    Comparison of regular convolutions (left), modulated deformable convolutions~\cite{Zhu19CVPR} (middle), and \OurConvName (right).
    Note that \OurConvName predicts a distinct spatiotemporal dilation for every point in the volume.
    Different colors indicate different modulation values.
    }
    \label{fig:conv_comparison}
\end{figure}

In parallel to the aforementioned developments in the video domain, another research area in computer vision was focusing on improving the performance of image-level segmentation networks.
To this end, one line of reasoning considers the limited receptive field of convolution operations as a drawback and aims to mitigate it. 
Even though a restricted receptive field is useful for weight sharing and imparting translation invariance, it is also a limitation for dense segmentation tasks where a wider view of the feature map can be beneficial.
Chen~\etal published a series of works~\cite{Chen17PAMI,Chen2017Arxiv,Chen2018ECCV} that use \emph{atrous convolutions} (also called \emph{dilated convolutions}) for semantic segmentation in images. Dilated convolutions effectively add padded zeros between the values in the convolutional kernel, thus enlarging the receptive field without incurring computational overhead or increasing the parameter count. Chen~\etal argued that the high degree of spatial downsampling usually applied in CNNs is detrimental for dense segmentation tasks. They instead maintained feature maps at a higher resolution and used dilated convolutions to capture a larger receptive field.

Another method for enhancing the receptive field of convolutions is the idea of \textit{deformable convolutions}~\cite{Dai17ICCV} (DCNv1). 
Here, the convolutional kernel can be arbitrarily shaped depending on the input feature map, as opposed to being a regular grid as in standard or dilated convolutions. Practically, this is realized by using the input feature map to predict offsets (or \emph{deformations}) to the sampling locations of the convolution operation. The underlying idea here is to enable the network to dynamically adapt the kernel based on the input image.
Zhu~\etal~\cite{Zhu19CVPR} further extended this by adding a dynamic \textit{modulation parameter} which scales the kernel weight value for each sampling location (DCNv2).
By simply using deformable convolutions as a drop-in replacement for standard convolutions, it was shown that the performance of a variety of network architectures could be improved for object detection and segmentation. 

Keeping these developments in mind gives rise to a question: Can 3D dilated/deformable convolutions retrace the success story of their 2D counter-parts and deliver improvements for video segmentation tasks? In this paper, we show that the answer is 'yes'. To this end, we propose a novel type of convolution called \OurConvName (\textbf{D}ynamic \textbf{D}ilated \textbf{3D} \textbf{Conv}olutions), which combines elements from dilated and deformable convolutions by dynamically learning a multiplicative scaling factor for the sampling locations of a convolutional kernel. 
Additionally, we also a predict a modulation parameter which dynamically scales the kernel weights based on the input features. 
We show that \OurConvName out-performs trivial extensions of dilated and deformable convolutions to 3D. Fig.~\ref{fig:conv_comparison} provides an illustrative comparison between: (i) standard 3D convolutions, (ii) a 3D extension of the modulated deformable convolutions proposed by Zhu~\etal~\cite{Zhu19CVPR}, (iii) our proposed \OurConvName.

In summary, our contributions are as follows: 
\begin{itemize}
    \item We propose a novel \OurConvName operator which can be used as drop-in replacements for standard convolutions in 3D CNNs to improve their performance on video segmentation tasks.
    \item We experimentally justify the efficacy of \OurConvName by applying it to two different 3D CNN based architectures~\cite{Athar20ECCV,Mahadevan20BMVC} and evaluating them on five different benchmarks~\cite{Perazzi16CVPR,Caelles19Arxiv,Yang19ICCV,Voigtlaender19CVPR,Qi21Arxiv}.
    \item We set a new state-of-the-art on the DAVIS 2016 Unsupervised challenge~\cite{Perazzi16CVPR} by achieving a $\JnF$ score of 86.0\%.
\end{itemize}

\section{Related Work}

\PAR{Image-level Segmentation:}Dense prediction tasks such as segmentation need to predict full resolution output maps and, at the same time, use multi-scale context for effective reasoning. Existing approaches for such tasks~\cite{Athar20ECCV, Chen17PAMI,Chen2017Arxiv,Chen2018ECCV, He17ICCV, Yu16ICLR} utilize dilated convolutions for this purpose, which dilate the convolutional kernel by a fixed factor to increase the receptive field, thus mitigating the need for down-sampling the image features.
Atrous Spatial Pyramid Pooling (ASPP)~\cite{Chen17PAMI} goes a step further by using multiple dilation rates on the same feature map to capture a multi-scale feature representation, and has been successfully used for both instance and semantic segmentation tasks \cite{Athar20ECCV, He17ICCV}.

Although dilated convolutions and ASPP help capture objects of different sizes, the convolutional kernels have fixed geometric structures since the dilation rates are constant. 
Several existing works~\cite{Dai17ICCV, Ding20CVPRW, Zhu19CVPR} attempt to adapt these kernels by learning offsets or other transformation parameters from the image features. Spatial Transformer Networks (STN)~\cite{Jaderberg15NIPS} learn deformations of the sampling grid, for a regular convolution operation, from the input feature map and warp the sampling grid based on the learnt deformation parameters.
Deformable Convolutional Network (DCNv1)~\cite{Dai17ICCV} on the other hand apply learned offsets to the sampling locations of a regular convolution, thereby enhancing its capability of capturing non-rigid transformations. DCNv1 can adapt to varying object sizes and scene geometry, and is shown to be effective for image-level tasks such as object detection and segmentation~\cite{Dai17ICCV}. Nevertheless, the sampling locations learned by DCNv1 often spread beyond the region of interest, which can lead to unnecessary feature influences. To overcome this issue,  Zhu~\etal introduced DCNv2~\cite{Zhu19CVPR} where, in addition to the offsets, a dynamic modulation parameter is learned which scales the kernel weights. This parameter gives the convolution kernels additional freedom to adjust the influence of the sampled regions. A teacher network based on R-CNN~\cite{Girshick14CVPR} is then used to train this modulation mechanism, where the teacher provides additional guidance to learn a more focused feature representation.

\OurConvName, similar to deformable convolutions~\cite{Dai17ICCV, Zhu19CVPR}, can be directly plugged-in to any existing architecture and improve its performance. However, unlike deformable convolutions, \OurConvName works with 3D models and can be used effectively for segmentation tasks in videos. In addition, the modulation mechanism used in \OurConvName does not need additional supervision from a teacher network as in DCNv2~\cite{Zhu19CVPR}.

\PAR{Video Processing using 3D Convolutions:}Videos can be interpreted as 3D data with the third dimension being time. To leverage temporal context effectively, video classification networks~\cite{Ji12PAMI,Karpathy14CVPR,Tran15ICCV,Varol17PAMI} successfully use 3D-CNNs and show their superior performance. However, unlike segmentation tasks, these networks do not need large resolution feature maps, and hence the increase in computational overhead caused by 3D-CNNs is still manageable. Recent works in the field of Unsupervised Video Object Segmentation~\cite{Hou19BMVC, Mahadevan20BMVC}, which target \textit{foreground-background} segmentation,  have also adapted 3D-CNNs to improve the segmentation performance. Hou~\etal~\cite{Hou19BMVC} uses an encoder-decoder architecture based on a variant of 3D-CNNs called R2plus1D~\cite{Tran18CVPR}, and insert an ASPP after the last layer of the encoder.
However, they adopt a relatively shallow network to compensate for the additional computation needed by ASPP, which in turn affects the final performance. Mahadevan~\etal~\cite{Mahadevan20BMVC} on the other hand employ a much deeper channel-separated 3D-CNN~\cite{Tran19ICCV} as backbone with much fewer parameters in combination with their novel 3D Global Convolutions and 3D Refinement modules in the decoder, and achieve state-of-the-art results. In this paper, we show that augmenting \cite{Mahadevan20BMVC} with \OurConvName further improves the network performance even with significantly less training data.

\PAR{Instance Segmentation in Videos:}Multi-instance Segmentation in Videos has recently emerged as a popular field due to its applicability in autonomous driving and robotics. Some of the popular tasks in this domain are Video Object Segmentation (VOS)~\cite{Caelles19Arxiv,Perazzi16CVPR}, Video Instance Segmentation (VIS)~\cite{Yang19ICCV}, and the more recent Occluded Video Instance Segmentation (OVIS)~\cite{Qi21Arxiv}. Here the primary goal is to segment all object instances in a video and associate them over time. For VIS and OVIS, there is an additional task of classifying the predicted tracks into one of the predefined object categories. Multi Object Tracking and Segmentation (MOTS)~\cite{Voigtlaender19CVPR} is another similar task that focuses on autonomous driving scenes and requires segmenting and tracking cars and pedestrians in these scenarios.

Popular methods~\cite{Yang19ICCV, Voigtlaender19CVPR, Zulfikar19CVPRW, Zeng19ICCV, Luiten18ACCV} that tackle the aforementioned tasks typically first generate image-level instance proposals, and then associate them either by using multiple cues such as optical flow and re-id, or by learning some kind of pixel affinity based on attention mechanism~\cite{Zulfikar19CVPRW, Luiten18ACCV}. Most of these methods employ 2D dilated convolutions in the backbone, and process each frame separately, thereby not effectively making use of the larger temporal context. Bertasius~\etal recently proposed MaskProp\cite{Bertasius20CVPR}, which modifies Mask R-CNN~\cite{He17ICCV} with a mask propagation branch to adapt it to videos
The features from the middle frame of an input video clip are aligned with the remaining frames using learnt spatial offsets similar to~\cite{Bertasius18ECCV}. Unlike \OurConvName, the mask propagation branch in MaskProp only operates on two frames, and it does not use learnt temporal dilation. STEm-Seg~\cite{Athar20ECCV} is another architecture that is relevant to our work. It processes an input video clip and generates spatio-temporal embeddings that can be directly clustered to obtain temporally consistent instance segmentation masks. STEm-Seg is a bottom-up approach, and uses a decoder comprising of entirely 3D convolutions for this purpose. In this paper, we show that by plugging in \OurConvName to just the decoder of STEm-Seg further improves its performance.

\section{Method}
Our proposed ~\OurConvName predicts a dilation scaling factor and a modulation value for every pixel in the input feature map.
Before explaining this in detail, we will first briefly recap the details of existing 2D deformable convolutions~\cite{Dai17ICCV,Zhu19CVPR} in Sec.~\ref{sec:recap_dcn}:

\subsection{Deformable Convolutions in 2D}
\label{sec:recap_dcn}

Let $\vect{X} \in \mathbb{R}^{H\times W}$ denote the feature map for an image with resolution $H\times W$ (we ignore the channel dimension for ease of notation).
Let $\vect{X}(\mathbf{p})$ denote the value of $\vect{X}$ at coordinates $\mathbf{p} = (p_y, p_x)$. 
Furthermore, let $\mathbf{W}$ denote the weights of a given convolutional kernel with $K$ entries, and let $\mathcal{S} \in \mathbb{R}^{K\times 2}$ denote the sampling region for the convolution. E.g. for a standard $3\times 3$ convolution, $K=9$, and the sampling region $\mathcal{S} = \{ (-1,-1), (-1,0), ..., (1,0), (1,1)\}$. 

In deformable convolutions~\cite{Dai17ICCV}, this sampling region is shifted by a set of offsets predicted for every point in $\vect{X}$ which we denote with $\Delta S \in \mathbb{R}^{H\times W\times K\times 2}$. Zhu~\etal~\cite{Zhu19CVPR} additionally also predict a set of modulation parameters $\mathbf{M} \in \mathbb{R}^{H\times W\times K}$.
If we let $\vect{Y}$ denote the feature map obtained after applying the deformable convolution, then the value of $\vect{Y}$ at coordinates $\mathbf{p_0}$ is calculated as follows:

\begin{equation}
\label{eq:defconv_2d}
\begin{split}
\textstyle
    \vect{Y}(\mathbf{p}_0) = \mathlarger{\sum}\limits_{\mathbf{p}_n \in \mathcal{S}}&   {\color{black} \mathbf{M}(\mathbf{p}_0, \mathbf{p}_n)} \cdot \mathbf{W}(\mathbf{p}_n) \cdot \\
     &\vect{X}(\mathbf{p}_0 + \mathbf{p}_n {\color{black}+ \Delta S(\mathbf{p}_0, \mathbf{p}_n)}) 
\end{split}
\end{equation}

Here, $\mathbf{M}(\mathbf{p}_0, \mathbf{p}_n)$ and $\Delta S(\mathbf{p}_0, \mathbf{p}_n)$ are used to denote the modulation value and sampling offset, respectively, predicted for point $\mathbf{p}_0$ in $\vect{X}$ at sampling location $\mathbf{p}_n$ in the kernel. Thus, deformable convolutions are able to dynamically attend to spatial locations outside of the fixed sampling region $\mathcal{S}$ which standard convolutions are bound to follow. 

Note that Eq.~\ref{eq:defconv_2d} is a generalization of the standard convolutions operation: if 
$\Delta S (\mathbf{p}_0, \mathbf{p}_n) = 0$ and $\mathbf{M}(\mathbf{p}_0, \mathbf{p}_n) = 1$, the deformable convolution reduces to a standard convolution.
The modulation parameters in $\mathbf{M}$ are sigmoid activated and thus lie in the range $[0,1]$, but the offsets in $\Delta S$ are unconstrained and may be fractional.
Therefore, bilinear interpolation is applied to sample the input feature map~\cite{Dai17ICCV,Zhu19CVPR}. Points outside the feature map are assumed to have a value of 0.

\subsection{$\textbf{D}^\textbf{2}$Conv3D~ -~ Dynamic Dilated Convolutions}
\label{sec:our_convs}

The fixed grid structure of convolutions imposes useful inductive bias for computer vision tasks due to the regular grid structure of rasterized images.
However, convolutional filters cannot adapt to changes in the underlying geometry of a scene.
By contrast, deformable convolutions can adapt to changes in scene geometry, but lose the inductive bias imposed by the fixed grid structure of a standard convolution kernel.

\begin{figure}
    \centering
    \includegraphics[width=\textwidth]{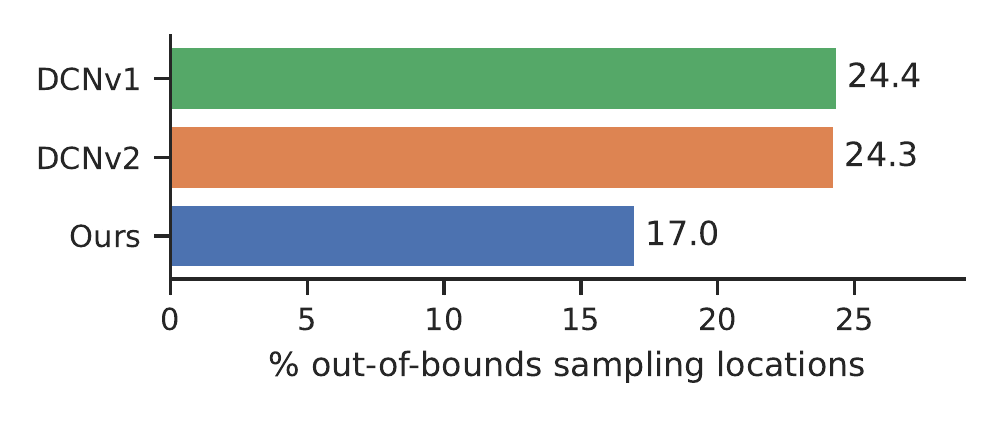}
    \caption{Average percentage of sampling locations outside the input volume during inference on DAVIS'16.}
    \label{fig:overall_oob}
\end{figure}

For video tasks, the size of the temporal dimension of an input spatio-temporal volume is usually orders of magnitude smaller than the spatial dimensions.
This means that a trivial extension of DCN~\cite{Dai17ICCV, Zhu19CVPR} to the 3D (spatio-temporal) domain often results in sampling locations that lie outside the spatio-temporal volume of the feature map. 
~\OurConvName on the other hand, strikes a compromise between these two types of convolutions: it maintains the grid structure of a convolutional kernel, but allows the kernel to be dilated dynamically and independently along each of the three dimensions.
Furthermore, compared to DCNv1 and DCNv2, sampling locations predicted by \OurConvName are better aligned with the input feature. This is shown in~\reffig{overall_oob}, where it can be seen that \OurConvName generates far less out of bounds sampling locations.
We refer to the supplementary material for further analysis of out-of-bounds sampling behaviour.

\PAR{Architecture.} In the 3D spatio-temporal domain of temporal dimension $T$, features can be redefined as $\vect{X} \in \mathbb{R}^{T\times H\times W}$, the sampling region of a convolution as $\mathcal{S} \in \mathbb{R}^{K\times 3}$, and the point coordinates $\mathbf{p} = (p_t,p_y,p_x)$. E.g. a $3\times 3\times 3$ convolution has $K=27$ entries.

\begin{figure}
    \centering
    \includegraphics[width=0.75\linewidth]{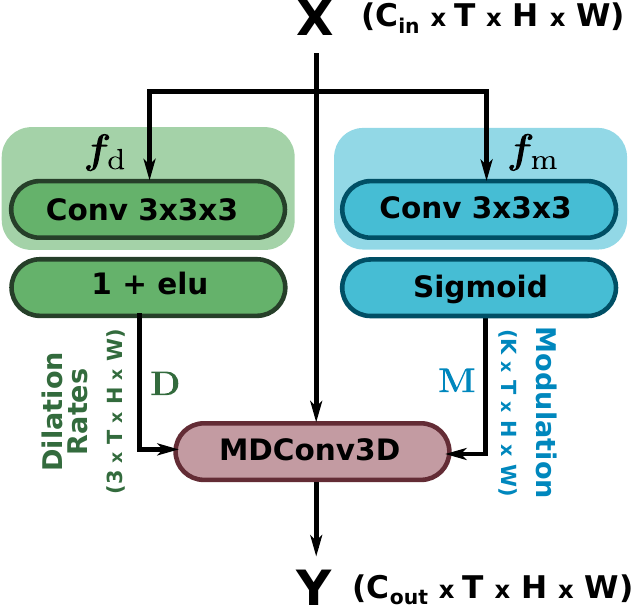}
    \caption{Architecture of the \OurConvName block. $f_d$ and $f_m$ are the names of the convolution layers which produce $\mathbf{D}$ and $\mathbf{M}$, respectively.}
    \label{fig:mdconv_block}
\end{figure}

Our proposed \OurConvName is a novel type of convolution that keeps the grid structure of normal convolutions while applying learnt dilations, and can be applied to spatio-temporal features. In contrast to deformable convolutions~\cite{Dai17ICCV,Zhu19CVPR} that learn $K$ additive offsets for the sampling region, we learn 3 multiplicative factors, one each for the $(t,y,x)$ dimensions, and apply them to the coordinates in the sampling region $\mathcal{S}$.
\OurConvName can thus be seen as dilated convolutions with dynamically learned dilation rates. We will henceforth use the term 'dilation map' to refer to the set of dilated rates predicted for $\vect{X}$, \ie $\vect{D} \in \mathbb{R}^{T\times H\times W\times 3}$.

To predict the dilation map, we input feature map $\vect{X}$ to a standard $3\times 3\times 3$ convolution $f_\text{d}$ followed by an elu activation function~\cite{Clevert15Arxiv} and addition by $1$:

\begin{equation}
    \mathbf{D} = 1 + \text{elu}(f_\text{d}(\vect{X}))
\label{eq:mdconv_dilation}
\end{equation}

This forces the values in $\mathbf{D}$ to lie in the range $[0, \infty)$. We found that simply applying a ReLU activation to $f_\text{d}(\vect{X})$ frequently results in zero gradients. By contrast, the formulation in Eq.~\ref{eq:mdconv_dilation} produces more well-behaved gradients during training, and also better results during inference (see Sec.~\ref{sec:ablation_activation_fn}).

Separately, a $3\times 3\times 3$ convolution $f_m$ is applied to $\vect{X}$ followed by sigmoid activation to produce the modulation map $\mathbf{M} \in \mathbb{R}^{T\times H\times W\times K}$ (\cf~\cite{Zhu19CVPR}). The value of the output feature map $\vect{Y}$ at point $\mathbf{p}_0$ with \OurConvName is then calculated as follows:

\begin{equation}
\label{eq:mdconv}
\begin{split}
\textstyle
    \vect{Y}(\mathbf{p}_0) = \mathlarger{\sum}\limits_{\mathbf{p}_n \in \mathcal{S}} &  {\color{black} \mathbf{M}(\mathbf{p}_0, \mathbf{p}_n)} \cdot \mathbf{W}(\mathbf{p}_n) \cdot \\
    & \vect{X}(\mathbf{p}_0 + (\mathbf{p}_n \cdot {\color{black} \mathbf{D}(\mathbf{p}_0)}))
\end{split}
\end{equation}

Here, with some abuse of notation, we use $\mathbf{p}_n \cdot \mathbf{D}(\mathbf{p}_0)$ to denote the multiplication of the sampling location coordinates $\mathbf{p}_n$ with the tuple of $3$ dilation rates in $\mathbf{D}$ at point $\mathbf{p}_0$.

\begin{figure*}[t]
    \centering
    \centering
    \rotatebox{90}{\resizebox{4.8cm}{!}{\parbox{7cm}{ \large \vspace{0px}\hspace{1pt} \textbf{Modulation} \hspace{6pt} \textbf{Dilations}  \hspace{11pt} \textbf{Results}}}}
    \includegraphics[width=0.48\textwidth]{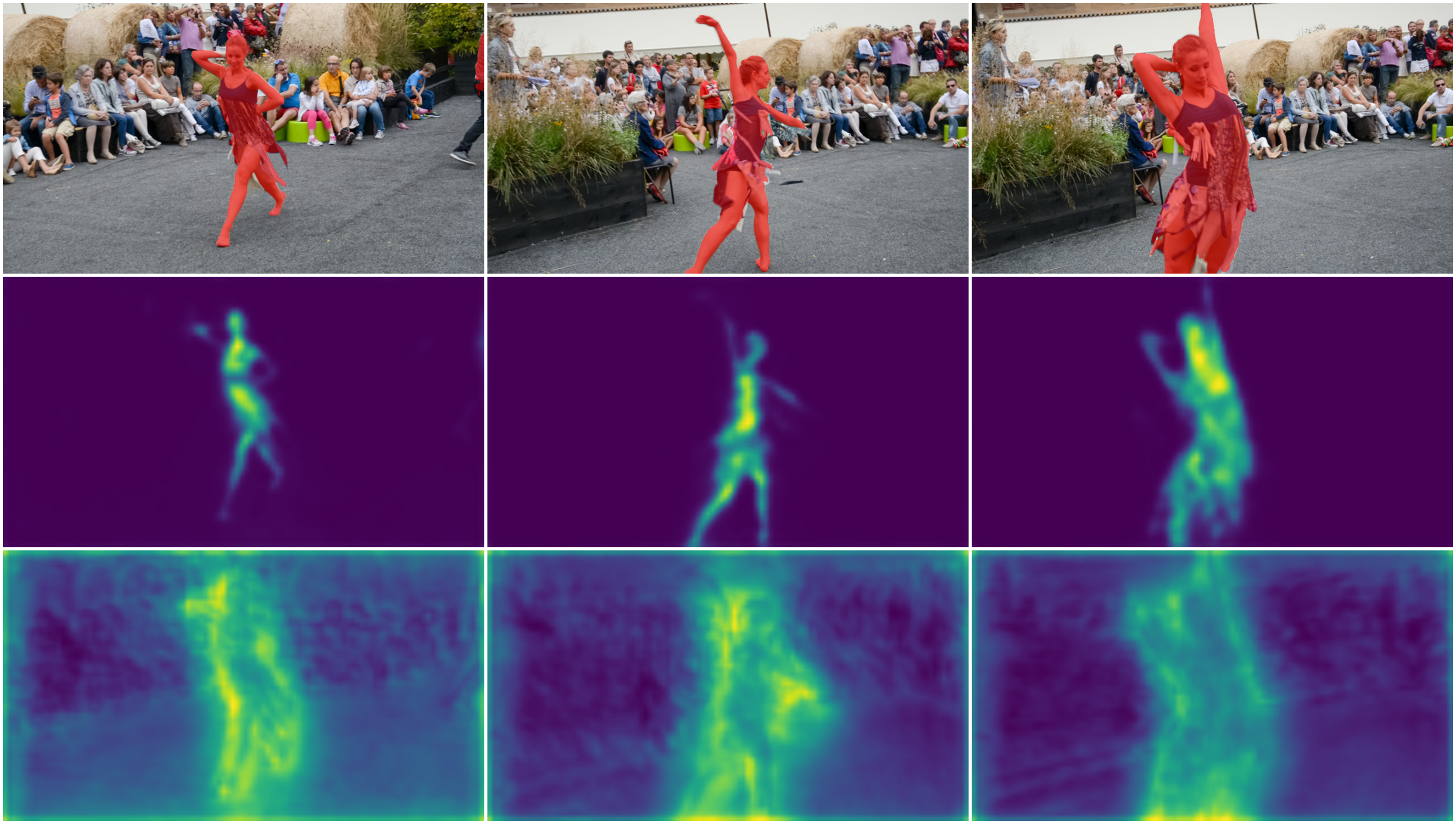}
    \unskip\ \vrule\unskip\ %
    \includegraphics[width=0.48\textwidth]{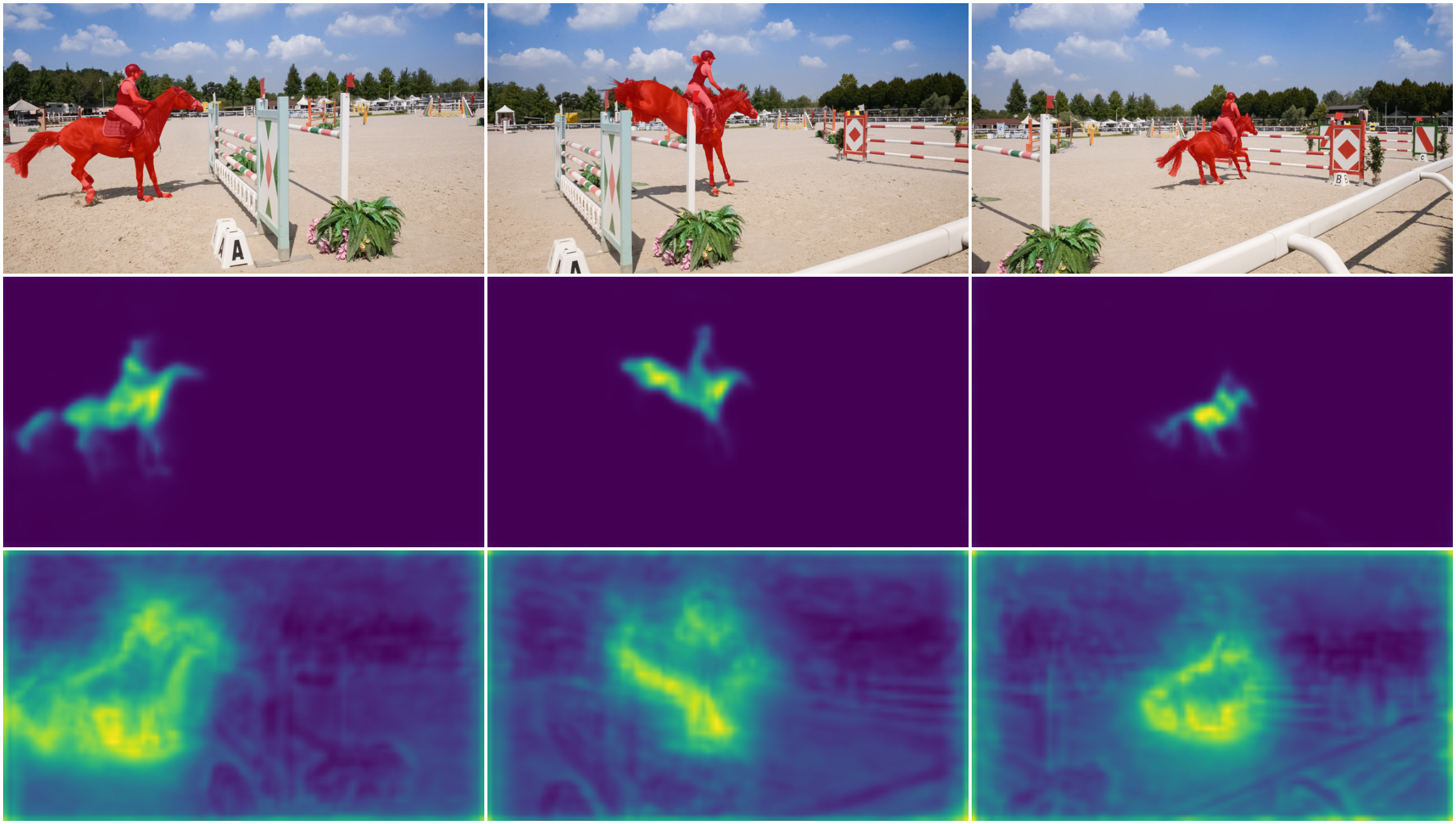}
    
    \caption{
    Qualitative results on DAVIS'16~\cite{Perazzi16CVPR}.
    Depicted are results from our model (top row), mean predicted dilation factors (middle row), and mean predicted modulation values.
    Lighter pixels denote higher values.
    Dilation factors are highest on the object, while modulation values are highest on the object boundary.
    Left sequence is \emph{dance-twirl}, right sequence is \emph{horsejump-high}.
    }
    \label{fig:qualitative_results}
\end{figure*}

Fig.~\ref{fig:mdconv_block} illustrates the architecture of a \OurConvName block, which comprises the actual \OurConvName layer, and also the two layers and activations required to produce $\mathbf{M}$ and $\mathbf{D}$. 
Note that, similar to Eq.~\ref{eq:defconv_2d}, \OurConvName is also a generalization of the convolution operation - if the modulation parameter and dilation rates are unity, \OurConvName reduces to a standard convolution. 
Moreover, it can also specialize to a 2D, 1D, or point-wise convolution by predicting one or more dilation rates as zero.
It is therefore possible to use an \OurConvName block as a drop-in replacement for 3D convolutions in existing pre-trained networks without modifying their behavior at the start of training. This can be done by simply copying the existing kernel weights to the \OurConvName layer, and initializing the weights and bias parameters in $f_\text{d}$ and $f_\text{m}$ with zeros.

\subsection{Qualitative Analysis}

\begin{figure*}[t]
    \centering
    \rotatebox{90}{\resizebox{4.8cm}{!}{\parbox{7cm}{ \large \vspace{0px}\hspace{1pt} \textbf{Modulation} \hspace{6pt} \textbf{Dilations}  \hspace{11pt} \textbf{Results}}}}
    \includegraphics[width=0.48\textwidth]{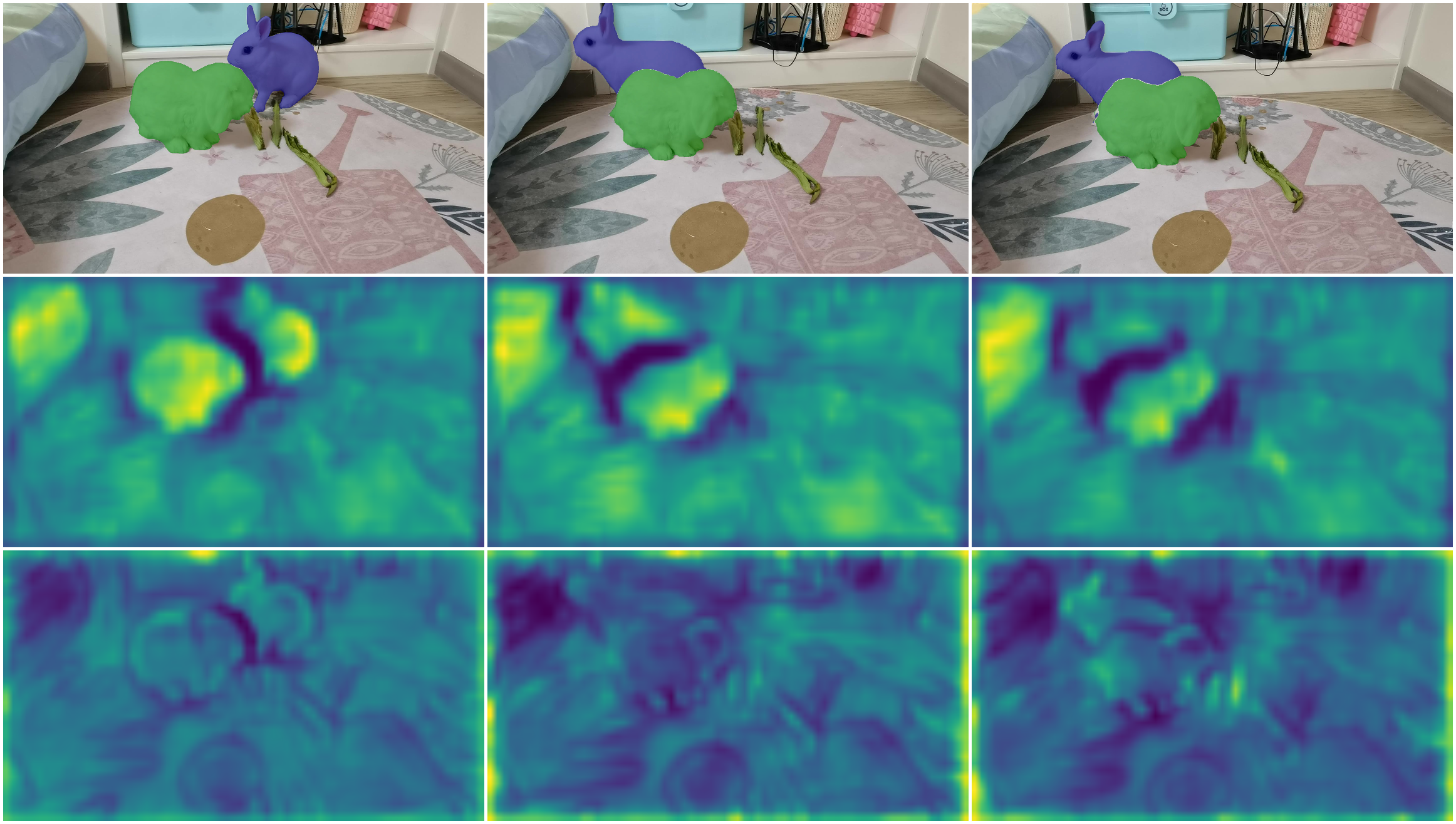}
    \unskip\ \vrule\unskip\ %
    \includegraphics[width=0.48\textwidth]{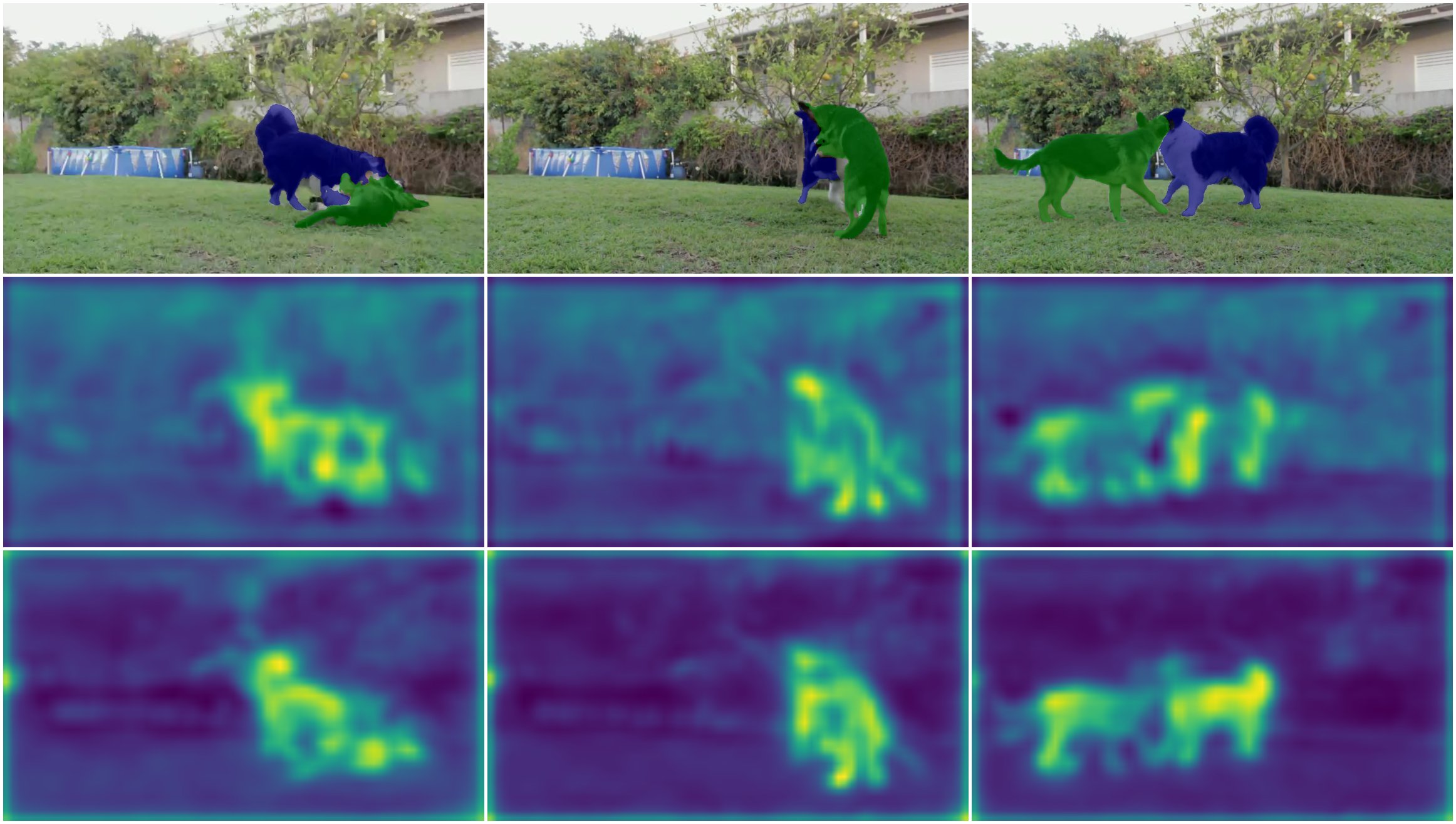}
    \caption{
    Qualitative results on OVIS~\cite{Qi21Arxiv}.
    Depicted are results from our model (top row), mean predicted dilation factors (middle row), and mean predicted modulation values.
    Left side: embedding decoder.
    Right side: semantic segmentation decoder.
    }
    \label{fig:qualitative_results_ovis}
\end{figure*}

The dilation rates and modulation values predicted by \OurConvName when it is used as an intermediate layer in a video object segmentation network are illustrated in \reffig{qualitative_results} and \reffig{qualitative_results_ovis}.
The visualizations shows the mean dilation rates (second row) and mean modulation values (third row).
It is evident that the network learns to use different dilation rates to distinguish the foreground object from the background.
Dilation rates are highest inside the object, medium on the object borders, and zero in the background.
The modulation values, on the other hand, are highest on the object boundaries.
This indicates that the network mostly focuses on refining the edges of the segmentation mask.

\section{Experiments on Video Saliency}

To justify the efficacy of \OurConvName, we apply it as a drop-in replacement for convolution layers in the 3D CNN architecture proposed by Mahadevan~\etal~\cite{Mahadevan20BMVC}. Subsequently, we conduct several ablations and also compare our results to existing state-of-the arts. The experimental evaluation is performed on the validation set of the DAVIS'16 Unsupervised Video Object Segmentation benchmark~\cite{Perazzi16CVPR}. The task here is to perform foreground segmentation of the 'salient' regions of the video. Here, saliency is subjectively defined as regions which undergo motion changes significant enough to capture the attention of the human eye. Note that even though the word 'unsupervised' occurs in the benchmark name, this is a fully supervised task where labeled training data is provided.

\subsection{Network Architecture}

We build on the baseline network architecture proposed by Mahadevan~\etal~\cite{Mahadevan20BMVC} which has a compact encoder-decoder architecture composed entirely of 3D convolutions. 
It accepts an input video clip and outputs a foreground probability map for each pixel in the input clip. 
The encoder is an efficient channel-separated 3D variant of ResNet-152~\cite{Tran19ICCV} from which feature maps are extracted at four different spatially downsampled scales (4x, 8x, 16x, 32x). The decoder is made up of three so-called \textit{Refinement modules} which upsample the given input feature map and combine it with the encoder feature map at the corresponding scale. 
To process arbitrarily long videos, the input video is split into multiple 8-frame clips with an overlap of 3 frames between successive clips. The output probability maps for the overlapping frames are subsequently averaged to get the final pixel foreground probabilities.
We refer the reader to \cite{Mahadevan20BMVC} for more details of the baseline.

\PAR{Our Changes.} We replace all convolution layers in the first two \textit{Refinement modules} with \OurConvName. These two modules are the ones which upsample the current feature map and combine it with the encoder features at the 16x and 8x spatially downsampled scales. We will henceforth refer to these modules as \texttt{rf1} and \texttt{rf2}.

Moreover, we apply GroupNorm~\cite{Wu18ECCV} to the outputs of the convolution layers in order to improve the gradient flow to deeper network layers.
It should be noted that \OurConvName is slower than a standard convolution, so it is infeasible to apply it to high resolution feature maps (\eg at the 4x spatial scale). We chose \texttt{rf1} and \texttt{rf2} as the replacement locations since this yields a good trade-off between performance and speed. We found that replacing layers in the encoder only provides a minor improvement that does not justify the increased computation/memory overhead.

\subsection{Training}

We initialize the encoder with weights from a publicly available model which is trained for video action classification on Kinetics400~\cite{Kay17Arxiv} and Sports-1M~\cite{Karpathy14CVPR}.
The weights of layers which predict offsets and modulation parameters are initialized to zero. All other weights in the decoder are initialized randomly. The network is trained for 20 epochs on 8-frame clips sampled randomly from the DAVIS'17~\cite{PontTuset17Arxiv} training set without any data augmentation.
We use the Lovasz-Hinge loss~\cite{Berman18CVPR} and train the network using using the Adam optimizer~\cite{Kingma14ICLR} with a learning rate of $10^{-5}$, which is reduced by a factor of 0.1 after 10 epochs.
Additionally, we employ gradient clipping by limiting the L2-norm of the gradient to a maximum value of 10.

\subsection{Ablations on Deformations}

We report the results of our first set of ablations in \reftab{offset_ablations}.
The most comparable baseline setting from \cite{Mahadevan20BMVC} which uses the same (pre-)training data as we do achieves a $\JnF$ score of 82.2.
For a fair comparison, we re-train this baseline with our training settings and added normalization layers. This 'Revised Baseline' achieves a $\JnF$ score of 83.5.

\PAR{ASPP.} Here, we replace the convolution layers in \texttt{rf1} and \texttt{rf2} with ASPP blocks~\cite{Chen17PAMI}, which comprise multiple dilated convolutions in parallel, but with a fixed dilation rate. Doing so actually reduces the $\JnF$ score from 83.5 to 82.6.

\PAR{Deformable Convolutions.} Next, we use 3D extensions of existing deformable convolutions in \texttt{rf1} and \texttt{rf2}. Using DCNv1~\cite{Dai17ICCV} improves the $\JnF$ by 0.9\% over the revised baseline (83.5 \textrightarrow ~84.4), while using DCNv2~\cite{Zhu19CVPR} provides a 1.3\% improvement (83.5 \textrightarrow ~84.8).
This clearly shows that dynamically adapting the convolution sampling locations is beneficial for video segmentation tasks.

\begin{table}[t]
    \centering
    \begin{tabular}{l|c|ccc}
        \toprule
        Variant                & Mod. & $\JnF$ & FPS & Mem. (GB) \\
        \midrule
        Baseline \cite{Mahadevan20BMVC} & -- & 82.2 & -- & -- \\
        Revised Baseline                & -- & 83.5 & 4.96 & 1.92\\
        \midrule
        ASPP                            & -- & 82.6 & 4.50 & 1.96 \\
        DCNv1~\cite{Dai17ICCV} & \xmark & 84.4 & 4.41 & 1.95 \\
        DCNv2~\cite{Zhu19CVPR} & \cmark & 84.8 & 4.36 & 1.96 \\
        \midrule
        (I)\phantom{I)} S  ($|\mathbf{D}|=1$)  & \cmark & 84.2 & 4.49 & 1.94 \\
        (II)\phantom{I} S ($|\mathbf{D}|=2$)  & \cmark & 84.3 & 4.46 & 1.94 \\
        (III) S+T ($|\mathbf{D}|=3$) & \xmark & 85.0 & 4.46 & 1.93 \\
        (IV) S+T ($|\mathbf{D}|=3$) & \cmark & \textbf{85.5} & 4.46 & 1.94 \\
        \bottomrule
    \end{tabular}
    \caption{
    Ablation on offset modality. Runtime and memory were measured during inference on an NVIDIA GTX1080ti.
    Mod: Modulation, S: spatial, S+T: spatio-temporal
    }
    \label{tab:offset_ablations}
\end{table}

\PAR{\OurConvName Variants.} 
Finally, we replace the convolutions in \texttt{rf1} and \texttt{rf2} with variants of dynamic dilated convolutions.
In (I), a single dilation rate is predicted for the two spatial dimensions, fixing the temporal dilation to 1.
This improves the $\JnF$ from 83.5 in the revised baseline to 84.2, but still lags behind DCNv1 and DCNv2. In (II) we allow different dilation rates for the two spatial dimensions, but the temporal dilation remains fixed. Doing so yields an insignificant improvement of 0.1\% $\JnF$ over variant (I).

In variant (III), we predict separate dilation rates for the 3 spatio-temporal dimensions, but do not predict modulation parameters. This variant achieves 85.0 $\JnF$ which is 1.5\% higher than the revised baseline (83.5), and 0.6\% higher than DCNv1 variant (84.4) which also does not use modulation parameters. Finally, variant (IV) is the full \OurConvName with modulation parameters (as explained in Sec.~\ref{sec:our_convs}). This achieves a $\JnF$ score of 85.5 which is 2\% higher than the revised baseline, and 0.7\% higher than the second-best performing DCNv2 variant. The fact that variant (IV) achieves a 1.8\% higher $\JnF$ compared to variant (II)
shows the effectiveness of the dynamic temporal dilation rate predicted by the network.

In light of these results, we can see that for video segmentation tasks, \OurConvName out-performs 3D extensions of existing deformable convolutions despite having fewer parameters (our dilation map $\mathbf{D}$ has 3 channels, whereas the offset maps for DCNv1 and DCNv2 have 81 channels for a $3\times 3\times 3$ convolution).
This justifies our earlier claim that \OurConvName allow the network to adapt to geometric scene variations in the input features while retaining the useful inductive bias associated with the grid shaped structure of a convolutional kernel.

\begin{table}[t]
    \centering
    \begin{tabular}{c|cccc}
        \toprule
        Activation & None & 1 + ReLU & ReLU & 1 + elu \\
        \midrule
        $\JnF$     & 83.8 & 83.9 & 85.1 & \textbf{85.5} \\
        \bottomrule
    \end{tabular}
    \caption{
    Ablations on dilation map activation function.
    }
    \label{tab:dilation_activation_ablations}
\end{table}

%
%

\subsection{Ablations on Dilation Activation Function}
\label{sec:ablation_activation_fn}

\reftab{dilation_activation_ablations} examines the effect of using different activation functions to predict the dilation map $\mathbf{D}$. Using no activation achieves a $\JnF$ score of 83.8. Note that this setting allows the network to predict a negative dilation rate, which mirrors the convolution kernel along that dimension.
'1 + ReLU' restricts the range of dilations to $[1,\infty)$. Looking at Fig.~\ref{fig:qualitative_results}, we see that the network often specializes \OurConvName to a point-wise convolution for background regions by predicting very small dilation rates (close to zero). Since '1 + ReLU' disallows such behavior, the $\JnF$ reduces by 1.2\% to 83.9 compared to the 85.1 achieved by just applying a ReLU. 
Finally, our chosen '1 + elu' activation (85.5 $\JnF$) out-performs ReLU (85.1 $\JnF$) by 0.4\% due to improved gradient flow for small dilation rates.

\subsection{Comparison with State of the Art}
\label{sec:sota_comparison}

\begin{table}[t]
    \centering
    \begin{tabular}{l|c|cc}
        \toprule
        Method & $\JnF$ & $\J$-mean & $\F$-mean \\
        \midrule
        3D-CNN~\cite{Hou19BMVC}           & 77.8 & 78.3 & 77.2 \\
        AGNN$^{\ast\ddagger}$~\cite{Wang19ICCV}            & 79.9 & 80.7 & 79.1 \\
        COSNet$^{\ddagger}$~\cite{Lu19CVPR}            & 80.0 & 80.5 & 79.5 \\
        STEm-Seg$^{\ast}$~\cite{Athar20ECCV}       & 80.6 & 80.6 & 80.6 \\
        ADNet$^{\ast\mathsection}$~\cite{Yang19ICCVAnchorDiff} & 81.1 & 81.7 & 80.5 \\
        MATNet$^{\dagger}$~\cite{Zhou20TIP}           & 81.6 & 82.4 & 80.7 \\
        DFNet$^{\ast\mathsection}$~\cite{Zhen20ECCV}           & 82.6 & 83.4 & 81.8 \\
        3DC-Seg$^{\ast\ast}$~\cite{Mahadevan20BMVC}    & 84.5 & 84.3 & 84.7 \\
        RTNet$^{\dagger\ddagger}$~\cite{Ren21CVPR}            & 85.2 & \textbf{85.6} & 84.7 \\
        \midrule
        Ours         & 85.5 & 84.9 & 86.0 \\
        Ours (dense) & \textbf{86.0} & 85.5 & \textbf{86.5} \\
        \bottomrule
    \end{tabular}
    \caption{
    Quantitative results on the DAVIS'16 unsupervised validation set.
    $^\dagger$Optical flow
    $^\ddagger$CRF post-processing
    $^\ast$Multi-scale inference
    $^{\ast\ast}$large-scale pretraining
    $^\mathsection$heuristic post-processing
    }
    \label{tab:sota_davis16}
\end{table}

\reftab{sota_davis16} shows the results of our method (baseline~\cite{Mahadevan20BMVC} with \OurConvName) in comparison to existing state-of-the-art approaches on the validation set of the DAVIS'16 Unsupervised Video Object Segmentation.
Our method achieves 85.5 $\JnF$ using a clip overlap of 3 frames during inference. If this overlap is increased to 7 frames, the $\JnF$ increases to 86.0 at the cost of slower run-time. 

Our method out-performs the current state-of-the-art RTNet~\cite{Ren21CVPR} (85.2 $\JnF$) by 0.8\% even though RTNet uses CRF post-processing and optical flow as an external cue. In fact, almost all other works either train on significantly more data~\cite{Athar20ECCV,Mahadevan20BMVC}, or use additional performance improvement cues such as optical flow~\cite{Ren21CVPR}, CRF postprocessing~\cite{Wang19ICCV,Lu19CVPR,Ren21CVPR} or other heuristics~\cite{Yang19ICCVAnchorDiff,Zhen20ECCV}. By contrast, we use only the DAVIS'17 dataset for training and use no other external cues, augmentations or post-processing techniques. 
Also note that our $\F$ score (86.5) is significantly higher than the second-highest (84.7), indicating that \OurConvName can produce highly accurate object boundaries.

\section{Multi-Instance Segmentation in Video}

To show the generalization capability of \OurConvName, we evaluate on four other benchmarks involving multi-instance segmentation in videos. 

\subsection{Network Architecture}
\label{sec:stemseg_arch}

We extend the network architecture of STEm-Seg~\cite{Athar20ECCV}, which is a single-stage approach that segments multiple object instances by clustering per-pixel embeddings in a given input video clip. STEm-Seg also has an encoder-decoder architecture, but unlike \cite{Mahadevan20BMVC}, it only has 3D convolutions in the decoder.
For our experiments, we replace the convolution layers in the two 'deepest' blocks of the decoder (which process the 32x and 16x down-sampled feature maps) with \OurConvName. Furthermore,  in order to reduce training time, we use a lighter ResNet-50 encoder backbone compared to the ResNet-101 used in the original paper~\cite{Athar20ECCV}. All other details, including the training schedule, are kept identical.

\begin{table*}[t]
    \centering
    \begin{tabular}{l|cccc|cccc}
        \toprule
               & \multicolumn{4}{c|}{Car}      & \multicolumn{4}{c}{Pedestrian} \\
        Conv Type & sMOTSA & MOTSA & MOTSP & IDS & sMOTSA & MOTSA & MOTSP & IDS \\
        \midrule
        Baseline   & 66.6 & 76.6 & 87.5 & 67 & 38.5 & 54.4 & \textbf{77.1} & 33 \\
        DCNv1     & 68.7 & 78.8 & \textbf{87.7} & 79 & 35.4 & 51.1 & 76.0 & 54 \\
        DCNv2     & 67.5 & 77.7 & 87.6 & 86 & 38.7 & 54.2 & 76.5 & \textbf{28} \\
        \OurConvName & \textbf{69.7} & \textbf{80.0} & 87.6 & \textbf{66} & \textbf{42.6} & \textbf{58.3} & 76.7 & 46 \\
        \bottomrule
    \end{tabular}
    \caption{Performance improvements on KITTI MOTS. Baseline is STEm-Seg~\cite{Athar20ECCV} with a ResNet50 backbone.}
    \label{tab:kitti_mots}
\end{table*}

\begin{table}[t]
    \centering
    \begin{tabular}{l|c|cc}
        \toprule
        Conv Type & $\JnF$ & $\J$-mean & $\F$-mean \\
        \midrule
        Baseline    & 63.4 & 60.3 & 66.5 \\
        DCNv1~\cite{Dai17ICCV}      & 63.2 & 59.7 & 66.6 \\
        DCNv2~\cite{Zhu19CVPR}      & \textbf{64.6} & \textbf{61.0} & 68.2 \\
        \OurConvName  & \textbf{64.6} & 60.8 & \textbf{68.5} \\
        \bottomrule
    \end{tabular}
    \caption{Performance improvements on DAVIS'19. Baseline is STEm-Seg~\cite{Athar20ECCV} with a ResNet50 backbone.}
    \label{tab:davis_inst}
\end{table}

\begin{table}[t]
    \centering
    \begin{adjustbox}{max width=\textwidth}
    \begin{tabular}{c|l|c|cccc}
        \toprule
        Dataset & Conv Type & mAP & AP50 & AP75 & AR1 & AR10 \\
        \midrule
        \multirow{4}{*}{YVIS} & Baseline    & 30.6 & 50.7 & 33.5 & 31.6 & 37.1 \\
         & DCNv1~\cite{Dai17ICCV} & 31.7 & 50.8 & 34.0 & 31.9 & 37.8 \\
         & DCNv2~\cite{Zhu19CVPR} & 29.4 & 48.1 & 31.8 & 30.4 & 36.1 \\
         & \OurConvName  & \textbf{32.3} & \textbf{51.3} & \textbf{34.7} & \textbf{32.2} & \textbf{38.1} \\
        \midrule
        \multirow{4}{*}{OVIS} & Baseline    & 14.3 & 31.5 & 12.4 & 10.2 & 20.7 \\
         & DCNv1~\cite{Dai17ICCV}         & \textbf{15.9} & \textbf{34.0} & 13.2 & \textbf{10.8} & \textbf{22.4} \\
         & DCNv2~\cite{Zhu19CVPR}         & 14.9 & 31.6 & \textbf{13.8} & 10.5 & 21.5 \\
         & \OurConvName  & 15.2 & 33.8 & 13.7 & 10.6 & 22.2 \\ 
        \bottomrule
    \end{tabular}
    \end{adjustbox}
    \caption{Performance improvements on YoutubeVIS (YVIS) and OVIS. Baseline is STEm-Seg~\cite{Athar20ECCV} with a ResNet50 backbone.}
    \label{tab:ytvis_ovis}
\end{table}

\subsection{Benchmarks}
We compare the performance of STEm-Seg enhanced with \OurConvName on four popular, challenging benchmarks for multi-instance segmentation in videos. These are briefly described below:

\PAR{KITTI-MOTS.}
KITTI-MOTS~\cite{Voigtlaender19CVPR} is an extension of the popular KITTI dataset for Multi-Object Tracking (MOT)~\cite{Geiger12CVPR} which requires pixel-precise object masks as opposed to bounding boxes, hence the name MOTS (Multi-Object Tracking and Segmentation).
It contains 21 lengthy videos captured from a moving vehicle wherein the task is to segment and track all \emph{car} and \emph{pedestrian} object instances.
Performance is primarily assessed using the 'sMOTSA' measure~\cite{Voigtlaender19CVPR}, which is an extension of the CLEAR MOT metrics~\cite{Bernardin08JIVP} to account for pixel-precise segmentation masks.

\PAR{DAVIS'19 Unsupervised.} 
The DAVIS 2019 Unsupervised Video Object Segmentation benchmark~\cite{Caelles19Arxiv} requires all salient objects in the video to be segmented and tracked over time. The training and validation sets comprise 60 and 30 videos, respectively.
Similar to the DAVIS'16 Unuspervised benchmark, the evaluation metrics here are the $\J$ and $\F$ scores, which are averaged into a single $\JnF$ metric. For this benchmark however, the $\JnF$ is computed separately for each object instance.

\PAR{YouTube-VIS.}
The YouTube Video Instance Segmentation dataset~\cite{Yang19ICCV} consists of 2,883 videos with a total of more than 130k object instances. Here, in addition to segmenting and tracking object instances over time, a class label (from one of 40 known classes) also has to be assigned to each predicted instance. The evaluation measure is mean Average Precision (mAP).

\PAR{OVIS.}
Occluded Video Instance Segmentation~\cite{Qi21Arxiv} comprises 5,233 videos with labeled masks for 25 known object classes. The dataset is similar to YouTube-VIS in that it also uses mean Average Precision (mAP) as the evaluation measure, but is more challenging since it comprises longer videos where objects undergo significant occlusion.

\subsection{Results}
\label{sec:stemseg_results}

We compare the results of STEm-Seg with \OurConvName against the original baseline~\cite{Athar20ECCV}, and also against the case where 3D extensions of DCNv1~\cite{Dai17ICCV} and DCNv2~\cite{Zhu19CVPR} are used instead of \OurConvName.
On all four benchmarks, using \OurConvName in the decoder improves the results over the baseline, whereas DCNv1 and DCNv2 perform inconsistently, sometimes even degrading performance.

On the MOTS task (\reftab{kitti_mots}), using \OurConvName leads to a significant performance increase.
For the \emph{car} class, DCNv1 and DCNv2 improve the sMOTSA score over the baseline by 2.1 and 0.9, respectively, whereas \OurConvName yields a more profound improvement of 3.1. 
For the \emph{pedestrian} class, \OurConvName improves the sMOTSA score over the baseline by 4.1 (38.5 \textrightarrow ~42.6), whereas DCNv1 actually reduces the sMOTSA by 3.1 (38.5 \textrightarrow ~35.4) and DCNv2 yields only a minor 0.2 improvement.

On DAVIS'19 (\reftab{davis_inst}), \OurConvName improves the $\JnF$ to 64.6, which is 1.2\% higher than the baseline (63.4).
The 3D extension of DCNv2 achieves similar performance, while DCNv1 performs slightly worse than the baseline.
Both \OurConvName and DCNv2 significantly improve the $\F$ measure, which indicates that modulating kernel weights improves the networks ability to accurately predict the contours of object instances.

On YoutubeVIS and OVIS (\reftab{ytvis_ovis}), \OurConvName improves over the baseline by 1.7\% and 0.9\% mAP, respectively.
This indicates that \OurConvName improves segmentation quality for complex scenes with several occluded objects.
DCNv1 improves result on YouTube-VIS by 1.1\% mAP and achieves an even greater improvement on OVIS, where it boosts the mAP from 14.3 to 15.9, outperforming \OurConvName by 0.7\%.
On the other hand, DCNv2 improves over the baseline on OVIS, but performs worse on Youtube-VIS.
We note that our training schedule for OVIS may be sub-optimal since OVIS was not evaluated in the original paper~\cite{Athar20ECCV} and we simply used the same training setup and hyper-parameters as for YouTube-VIS.

\section{Conclusion}
In this paper, we presented \OurConvName, a new type of dynamic 3D convolution, and justified its efficacy by applying it to two different network architectures and five different video segmentation tasks.
Furthermore, we showed that \OurConvName out-performs 3D extensions of existing deformable convolutions because it experiences fewer out-of-bounds sampling locations and preserves the useful inductive bias associated with rectangular convolutional kernels.
Future work could explore shifting the kernel center position, and additional types of kernel shape deformation.

\PAR{Acknowledgements.}
This project was funded, in parts, by ERC Consolidator Grant DeeVise (ERC-2017-COG-773161), EU project CROWDBOT (H2020-ICT-2017-779942). Computing resources for several experiments were granted by RWTH Aachen University under project 'thes0863'. We thank Paul Voigtlaender, Istv{\'a}n S{\'a}r{\'a}ndi, Jonas Schult and Alexey Nekrasov for helpful discussions.

\clearpage

{\small
\bibliographystyle{ieee_fullname}
\bibliography{abbrev_short,egbib}

\begin{thebibliography}{10}\itemsep=-1pt

\bibitem{Athar20ECCV}
Ali Athar, Sabarinath Mahadevan, Aljo{\v{s}}a O{\v{s}}ep, Laura Leal-Taix{\'e},
  and Bastian Leibe.
\newblock Stem-seg: Spatio-temporal embeddings for instance segmentation in
  videos.
\newblock In {\em ECCV}, 2020.

\bibitem{Berman18CVPR}
Maxim Berman, Amal~Rannen Triki, and Matthew~B Blaschko.
\newblock The lov{\'a}sz-softmax loss: A tractable surrogate for the
  optimization of the intersection-over-union measure in neural networks.
\newblock In {\em CVPR}, 2018.

\bibitem{Bernardin08JIVP}
Keni Bernardin and Rainer Stiefelhagen.
\newblock Evaluating multiple object tracking performance: the clear mot
  metrics.
\newblock {\em JIVP}, 2008.

\bibitem{Bertasius20CVPR}
Gedas Bertasius and L. Torresani.
\newblock Classifying, segmenting, and tracking object instances in video with
  mask propagation.
\newblock In {\em CVPR}, 2020.

\bibitem{Bertasius18ECCV}
Gedas Bertasius, L. Torresani, and Jianbo Shi.
\newblock Object detection in video with spatiotemporal sampling networks.
\newblock In {\em ECCV}, 2018.

\bibitem{Caelles19Arxiv}
Sergi Caelles, Jordi Pont-Tuset, Federico Perazzi, Alberto Montes,
  Kevis-Kokitsi Maninis, and Luc {Van Gool}.
\newblock The 2019 davis challenge on vos: Unsupervised multi-object
  segmentation.
\newblock {\em arXiv}, 2019.

\bibitem{Cao20ECCV}
Jiale Cao, Rao~Muhammad Anwer, Hisham Cholakkal, Fahad~Shahbaz Khan, Yanwei
  Pang, and Ling Shao.
\newblock Sipmask: Spatial information preservation for fast image and video
  instance segmentation.
\newblock In {\em ECCV}, 2020.

\bibitem{Chen17PAMI}
Liang-Chieh Chen, George Papandreou, Iasonas Kokkinos, Kevin Murphy, and Alan~L
  Yuille.
\newblock Deeplab: Semantic image segmentation with deep convolutional nets,
  atrous convolution, and fully connected crfs.
\newblock {\em PAMI}, 2017.

\bibitem{Chen2017Arxiv}
Liang-Chieh Chen, George Papandreou, Florian Schroff, and Hartwig Adam.
\newblock Rethinking atrous convolution for semantic image segmentation.
\newblock {\em arXiv preprint arXiv:1706.05587}, 2017.

\bibitem{Chen2018ECCV}
Liang-Chieh Chen, Yukun Zhu, George Papandreou, Florian Schroff, and Hartwig
  Adam.
\newblock Encoder-decoder with atrous separable convolution for semantic image
  segmentation.
\newblock In {\em ECCV}, 2018.

\bibitem{Cho19CVPRW}
Donghyeon Cho, Sungeun Hong, Sungil Kang, and Jiwon Kim.
\newblock Key instance selection for unsupervised video object segmentation.
\newblock {\em CVPR Workshops}, 2019.

\bibitem{Clevert15Arxiv}
Djork-Arn{\'e} Clevert, Thomas Unterthiner, and Sepp Hochreiter.
\newblock Fast and accurate deep network learning by exponential linear units
  (elus).
\newblock {\em arXiv}, 2015.

\bibitem{Dai17ICCV}
Jifeng Dai, Haozhi Qi, Yuwen Xiong, Yi Li, Guodong Zhang, Han Hu, and Yichen
  Wei.
\newblock Deformable convolutional networks.
\newblock In {\em ICCV}, 2017.

\bibitem{Ding20CVPRW}
Mingyu Ding, Yuqi Huo, Hongwei Yi, Zhe Wang, Jianping Shi, Zhiwu Lu, and P.
  Luo.
\newblock Learning depth-guided convolutions for monocular 3d object detection.
\newblock In {\em CVPR Workshops}, 2020.

\bibitem{Fu21AAAI}
Yang Fu, Linjie Yang, Ding Liu, Thomas~S Huang, and Humphrey Shi.
\newblock Compfeat: Comprehensive feature aggregation for video instance
  segmentation.
\newblock {\em AAAI}, 2021.

\bibitem{Geiger12CVPR}
Andreas Geiger, Philip Lenz, and Raquel Urtasun.
\newblock Are we ready for autonomous driving? the kitti vision benchmark
  suite.
\newblock In {\em CVPR}, 2012.

\bibitem{Girshick14CVPR}
Ross~B. Girshick, Jeff Donahue, Trevor Darrell, and J. Malik.
\newblock Rich feature hierarchies for accurate object detection and semantic
  segmentation.
\newblock In {\em CVPR}, 2014.

\bibitem{He17ICCV}
K. He, G. Gkioxari, P. Doll{\'a}r, and R. Girshick.
\newblock Mask {R-CNN}.
\newblock In {\em ICCV}, 2017.

\bibitem{Hou19BMVC}
Rui Hou, Chen Chen, Rahul Sukthankar, and Mubarak Shah.
\newblock An efficient 3d {CNN} for action/object segmentation in video.
\newblock In {\em BMVC}, 2019.

\bibitem{Jaderberg15NIPS}
Max Jaderberg, Karen Simonyan, Andrew Zisserman, and koray kavukcuoglu.
\newblock Spatial transformer networks.
\newblock In {\em NIPS}, 2015.

\bibitem{Ji12PAMI}
Shuiwang Ji, Wei Xu, Ming Yang, and Kai Yu.
\newblock 3d convolutional neural networks for human action recognition.
\newblock {\em PAMI}, 2012.

\bibitem{Karpathy14CVPR}
Andrej Karpathy, George Toderici, Sanketh Shetty, Thomas Leung, Rahul
  Sukthankar, and Li Fei-Fei.
\newblock Large-scale video classification with convolutional neural networks.
\newblock In {\em CVPR}, 2014.

\bibitem{Kay17Arxiv}
Will Kay, Joao Carreira, Karen Simonyan, Brian Zhang, Chloe Hillier, Sudheendra
  Vijayanarasimhan, Fabio Viola, Tim Green, Trevor Back, Paul Natsev, et~al.
\newblock The kinetics human action video dataset.
\newblock {\em arXiv}, 2017.

\bibitem{Kingma14ICLR}
Diederik~P Kingma and Jimmy Ba.
\newblock Adam: A method for stochastic optimization.
\newblock {\em ICLR}, 2015.

\bibitem{Lu19CVPR}
Xiankai Lu, Wenguan Wang, Chao Ma, Jianbing Shen, Ling Shao, and Fatih Porikli.
\newblock See more, know more: Unsupervised video object segmentation with
  co-attention siamese networks.
\newblock In {\em CVPR}, 2019.

\bibitem{Luiten21IJCV}
Jonathon Luiten, Aljosa Osep, Patrick Dendorfer, Philip Torr, Andreas Geiger,
  Laura Leal-Taix{\'e}, and Bastian Leibe.
\newblock Hota: A higher order metric for evaluating multi-object tracking.
\newblock {\em IJCV}, 2021.

\bibitem{Luiten18ACCV}
Jonathon Luiten, Paul Voigtlaender, and Bastian Leibe.
\newblock Premvos: Proposal-generation, refinement and merging for video object
  segmentation.
\newblock In {\em ACCV}, 2018.

\bibitem{Mahadevan20BMVC}
Sabarinath Mahadevan, Ali Athar, Aljo{\v{s}}a O{\v{s}}ep, Sebastian Hennen,
  Laura Leal-Taix{\'e}, and Bastian Leibe.
\newblock Making a case for 3d convolutions for object segmentation in videos.
\newblock {\em BMVC}, 2020.

\bibitem{Perazzi16CVPR}
Federico Perazzi, Jordi Pont-Tuset, Brian McWilliams, Luc~Van Gool, Markus
  Gross, and Alexander Sorkine-Hornung.
\newblock A benchmark dataset and evaluation methodology for video object
  segmentation.
\newblock In {\em CVPR}, 2016.

\bibitem{Pinheiro16ECCV}
Pedro~O Pinheiro, Tsung-Yi Lin, Ronan Collobert, and Piotr Doll{\'a}r.
\newblock Learning to refine object segments.
\newblock In {\em ECCV}, 2016.

\bibitem{PontTuset17Arxiv}
Jordi Pont-Tuset, Federico Perazzi, Sergi Caelles, Pablo Arbel\'aez, Alexander
  Sorkine-Hornung, and Luc {Van Gool}.
\newblock The 2017 davis challenge on video object segmentation.
\newblock {\em arXiv}, 2017.

\bibitem{Qi21Arxiv}
Jiyang Qi, Yan Gao, Yao Hu, Xinggang Wang, Xiaoyu Liu, Xiang Bai, Serge
  Belongie, Alan Yuille, Philip~HS Torr, and Song Bai.
\newblock Occluded video instance segmentation.
\newblock {\em arXiv preprint arXiv:2102.01558}, 2021.

\bibitem{Ren21CVPR}
Sucheng Ren, Wenxi Liu, Yongtuo Liu, Haoxin Chen, Guoqiang Han, and Shengfeng
  He.
\newblock Reciprocal transformations for unsupervised video object
  segmentation.
\newblock In {\em CVPR}, 2021.

\bibitem{Tran15ICCV}
Du Tran, Lubomir Bourdev, Rob Fergus, Lorenzo Torresani, and Manohar Paluri.
\newblock Learning spatiotemporal features with 3d convolutional networks.
\newblock In {\em ICCV}, 2015.

\bibitem{Tran19ICCV}
Du Tran, Heng Wang, Lorenzo Torresani, and Matt Feiszli.
\newblock Video classification with channel-separated convolutional networks.
\newblock In {\em ICCV}, 2019.

\bibitem{Tran18CVPR}
Du Tran, Heng Wang, Lorenzo Torresani, Jamie Ray, Yann LeCun, and Manohar
  Paluri.
\newblock A closer look at spatiotemporal convolutions for action recognition.
\newblock In {\em CVPR}, 2018.

\bibitem{Varol17PAMI}
G{\"u}l Varol, Ivan Laptev, and Cordelia Schmid.
\newblock Long-term temporal convolutions for action recognition.
\newblock {\em PAMI}, 2017.

\bibitem{Ventura19CVPR}
Carles Ventura, Miriam Bellver, Andreu Girbau, Amaia Salvador, Ferran Marques,
  and Xavier Giro-i Nieto.
\newblock Rvos: End-to-end recurrent network for video object segmentation.
\newblock In {\em CVPR}, 2019.

\bibitem{Voigtlaender19CVPR_FEELVOS}
Paul Voigtlaender, Yuning Chai, Florian Schroff, Hartwig Adam, Bastian Leibe,
  and Liang-Chieh Chen.
\newblock Feelvos: Fast end-to-end embedding learning for video object
  segmentation.
\newblock In {\em CVPR}, 2019.

\bibitem{Voigtlaender19CVPR}
Paul Voigtlaender, Michael Krause, Aljosa Osep, Jonathon Luiten, Berin
  Balachandar~Gnana Sekar, Andreas Geiger, and Bastian Leibe.
\newblock Mots: Multi-object tracking and segmentation.
\newblock In {\em CVPR}, 2019.

\bibitem{Wang19ICCV}
Wenguan Wang, Xiankai Lu, Jianbing Shen, David~J Crandall, and Ling Shao.
\newblock Zero-shot video object segmentation via attentive graph neural
  networks.
\newblock In {\em ICCV}, 2019.

\bibitem{Wang21CVPR}
Yuqing Wang, Zhaoliang Xu, Xinlong Wang, Chunhua Shen, Baoshan Cheng, Hao Shen,
  and Huaxia Xia.
\newblock End-to-end video instance segmentation with transformers.
\newblock In {\em CVPR}, 2021.

\bibitem{Wojke17ICIP}
Nicolai Wojke, Alex Bewley, and Dietrich Paulus.
\newblock Simple online and realtime tracking with a deep association metric.
\newblock In {\em ICIP}, 2017.

\bibitem{Wu18ECCV}
Yuxin Wu and Kaiming He.
\newblock Group normalization.
\newblock In {\em ECCV}, 2018.

\bibitem{Yang19ICCV}
Linjie Yang, Yuchen Fan, and Ning Xu.
\newblock Video instance segmentation.
\newblock In {\em ICCV}, 2019.

\bibitem{Yang18CVPR}
Linjie Yang, Yanran Wang, Xuehan Xiong, Jianchao Yang, and Aggelos~K.
  Katsaggelos.
\newblock Efficient video object segmentation via network modulation.
\newblock In {\em CVPR}, 2018.

\bibitem{Yang21Arxiv}
Shusheng Yang, Yuxin Fang, Xinggang Wang, Yu Li, Chen Fang, Ying Shan, Bin
  Feng, and Wenyu Liu.
\newblock Crossover learning for fast online video instance segmentation.
\newblock {\em arXiv preprint arXiv:2104.05970}, 2021.

\bibitem{Yang19ICCVAnchorDiff}
Zhao Yang, Qiang Wang, Luca Bertinetto, Weiming Hu, Song Bai, and Philip H.~S.
  Torr.
\newblock Anchor diffusion for unsupervised video object segmentation.
\newblock In {\em ICCV}, 2019.

\bibitem{Yu16ICLR}
Fisher Yu and Vladlen Koltun.
\newblock Multi-scale context aggregation by dilated convolutions.
\newblock In {\em International Conference on Learning Representations}, 2015.

\bibitem{Zeng19ICCV}
Xiaohui Zeng, Renjie Liao, Li Gu, Yuwen Xiong, Sanja Fidler, and Raquel
  Urtasun.
\newblock Dmm-net: Differentiable mask-matching network for video object
  segmentation.
\newblock In {\em ICCV}, 2019.

\bibitem{Zhen20ECCV}
Mingmin Zhen, Shiwei Li, Lei Zhou, Jiaxiang Shang, Haoan Feng, Tian Fang, and
  Long Quan.
\newblock Learning discriminative feature with crf for unsupervised video
  object segmentation.
\newblock In {\em ECCV}, 2020.

\bibitem{Zhou20TIP}
Tianfei Zhou, Jianwu Li, Shunzhou Wang, Ran Tao, and Jianbing Shen.
\newblock Matnet: Motion-attentive transition network for zero-shot video
  object segmentation.
\newblock {\em TIP}, 2020.

\bibitem{Zhu19CVPR}
Xizhou Zhu, Han Hu, Stephen Lin, and Jifeng Dai.
\newblock Deformable convnets v2: More deformable, better results.
\newblock In {\em CVPR}, 2019.

\bibitem{Zulfikar19CVPRW}
I.~E. Zulfikar, J. Luiten, and B. Leibe.
\newblock {UnOVOST: Unsupervised Offline Video Object Segmentation and Tracking
  for the 2019 Unsupervised DAVIS Challenge}.
\newblock {\em CVPR Workshops}, 2019.

\end{thebibliography}
}

\clearpage
\twocolumn[{%
 \centering
 \Large \textbf{$\textbf{D}^\textbf{2} \text{Conv3D}$: Dynamic Dilated Convolutions for Object Segmentation in Videos \\ Supplementary Material \\[1cm]}
}]
\setcounter{equation}{0}
\setcounter{figure}{0}
\setcounter{table}{0}
\setcounter{page}{1}
\setcounter{section}{0}
\makeatletter
\renewcommand{\theequation}{S\arabic{equation}}
\renewcommand{\thefigure}{S\arabic{figure}}
\renewcommand{\thetable}{S\arabic{table}}
\renewcommand{\thesection}{S\arabic{section}}

\section{Modulation Map Visualization}
\begin{figure*}
    \centering
    \includegraphics[width=\textwidth]{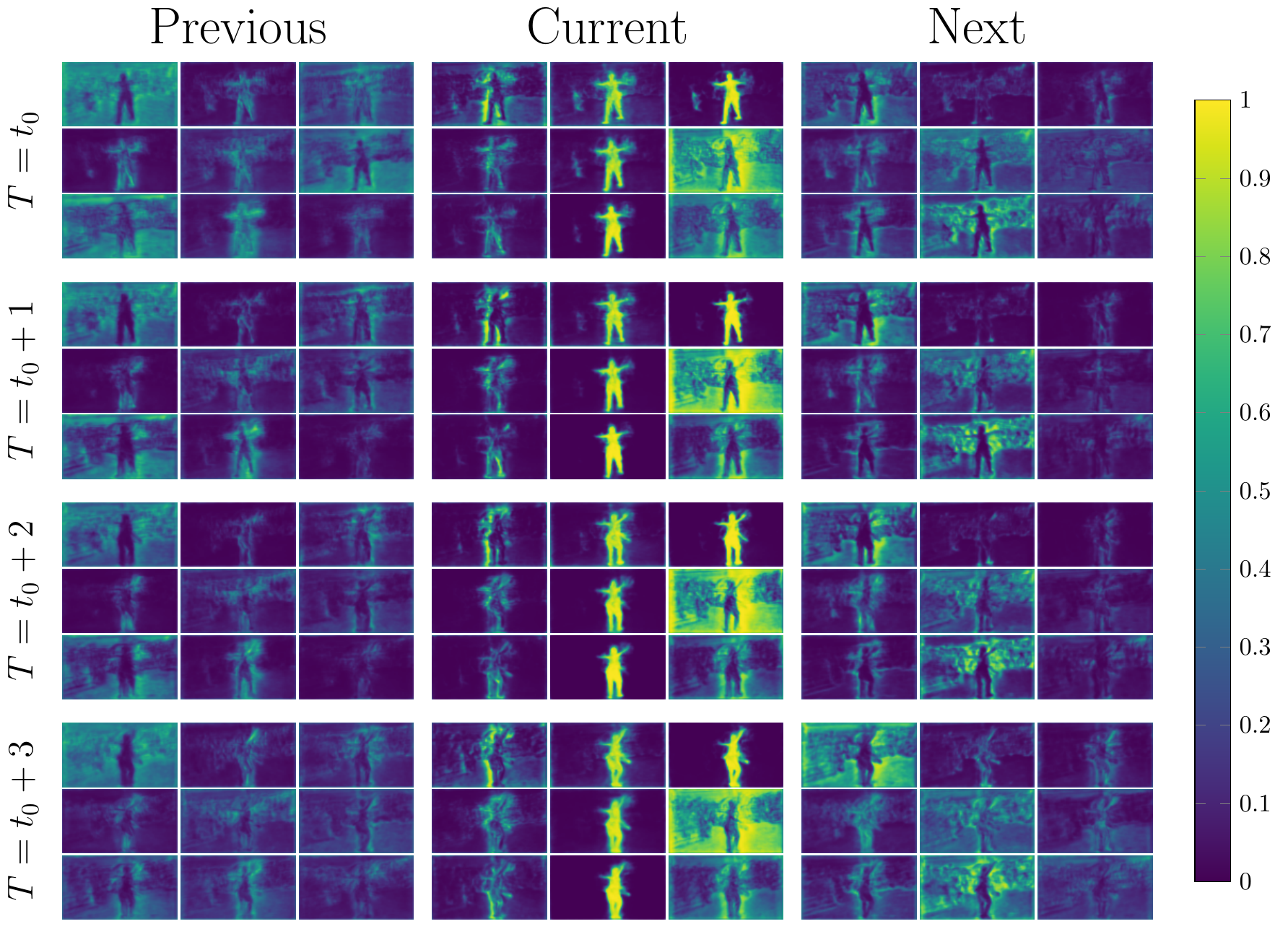}
    \caption{
    Modulation values predicted during inference on the \textit{dance-twirl} sequence in DAVIS'16. Recall that for a $3\times 3\times 3$ convolution, the modulation map $\textbf{M} \in \mathbb{R}^{T\times H\times W\times K}$ has $K=27$ channels for each pixel in the input feature map. Here we visualize these 27 channels by splitting them into a row of 3 image blocks, with each block having size $3\times 3$. Consider the row of image blocks for $T=t_0$ : here the image block under "Previous" corresponds to the modulation values predicted for those kernel weights which will be applied to the video features in the previous timestep ($T= t_0 - 1$). Likewise, "Curent" and "Next" show the modulation values for the kernel weights which will be applied to the video features from the current ($T=t_0$) and next ($T=t_0+1$) timesteps, respectively. The modulation map $\mathbf{M}$ is shown here for a total of 4 time-steps ($t_0, ..., t_0+3$); thus, there are four sets of image blocks along the vertical dimension.
    }
    \label{fig:all_modulation_values}
\end{figure*}

In \reffig{all_modulation_values}, we visualize a full volume of predicted modulation values for each of the convolutional layers in the refinement modules when \OurConvName is applied to~\cite{Mahadevan20BMVC}.
It is visible that every channel reacts to different parts of the foreground or background.
Kernel points that potentially sample from neighbouring frames receive higher modulation values on the object boundaries or on the background.
Kernel points that sample the current frame, however, have low modulation values in the background and larger modulation values on the object.

\section{Out-of-bounds Sampling Behaviour}
\label{out_of_bounds_sampling}
\begin{figure}[t]
    \centering
    \includegraphics[width=0.99\linewidth]{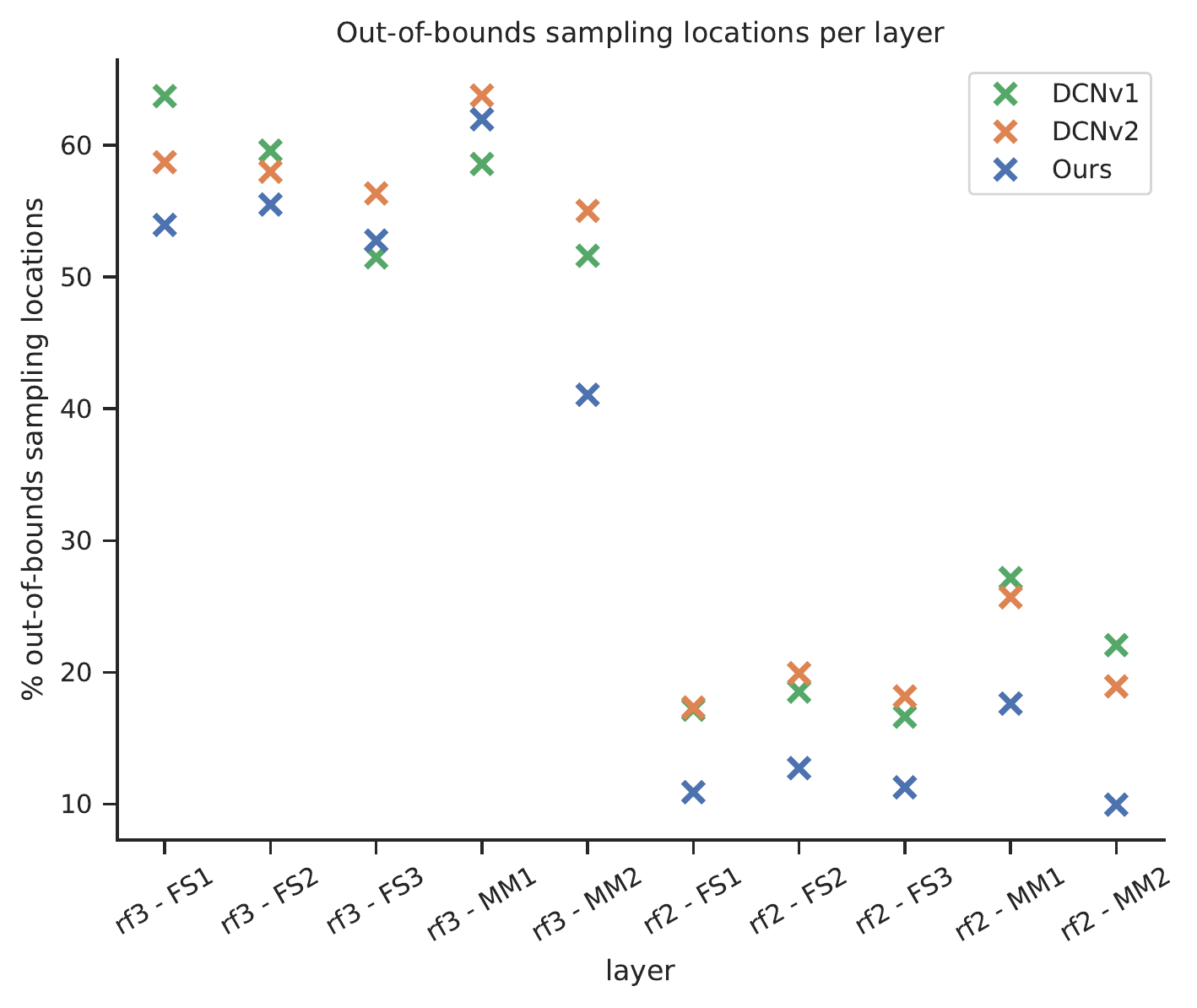}
    \caption{
    Percentage of out-of-bounds sampling locations, per layer.
    Measured during inference on DAVIS'16.
    }
    \label{fig:detailed_oobs}
\end{figure}

As mentioned in Sec. 3 of the main paper, we perform a detailed comparison of the percentage of sampling locations that are sampled outside the input feature volume, per convolutional layer, in \reffig{detailed_oobs}.
It can be observed that \OurConvName predicts fewer sampling locations beyond the input features than DCNv1 or DCNv2 in most of the cases.

\section{Runtime}
Although deformable convolutions are not as heavily optimized as regular convolutions, the impact on the runtime is small because we use them only on low-resolution feature maps.
Detailed runtimes can be found in \reftab{runtimes}

\begin{table}[t]
    \centering
    \begin{tabular}{l|cc}
        \toprule
        Model & \#Params (M) & Time (s/frame) \\
        \midrule
        AGNN~\cite{Wang19ICCV}            & 82.3 & 2.96 \\
        CosNet~\cite{Lu19CVPR}            & 81.2 & 0.45 \\
        STEm-Seg~\cite{Athar20ECCV}       & 50.1 & 1.42 \\
        ADNet~\cite{Yang19ICCVAnchorDiff} & 79.3 & 2.94 \\
        MatNet~\cite{Zhou20TIP}           & 142.7 & 0.75$^*$ \\
        DFNet~\cite{Zhen20ECCV}           & 64.7 & 0.28 \\
        3DC-Seg~\cite{Mahadevan20BMVC}    & 74.2 & 0.16 \\
        RTNet~\cite{Ren21CVPR}            & 277.2 & 0.29$^\dagger$ \\
        \midrule
        Revised Baseline                  & 74.2 & 0.2 \\
        Ours                              & 77.1 & 0.22 \\
        Ours (dense)                      & 77.1 & 1.07 \\
        \bottomrule
    \end{tabular}
    \caption{
    Runtimes during inference on DAVIS'16.
    Measured on an Nvidia GTX-1080Ti.
    $^\dagger$Not including time for CRF post-processing.
    $^*$ runtime reported on an Nvidia RTX-2080Ti.}
    \label{tab:runtimes}
\end{table}

\section{Comparison with State-of-the-art}

\PAR{DAVIS 2019:} Table.~\ref{tab:davis19-bench} reports the results of the state-of-the-art methods on DAVIS'19 unsupervised validation set. The methods that are grayed out do not use 3D convolutions and hence \OurConvName cannot be plugged-in to them for a direct comparison. UnOVOST\,\cite{Zulfikar19CVPRW} performs the best among all the methods with a $\mathcal{J} \& \mathcal{F}$ score of $67.0\%$, but it uses multiple 2D networks along with heuristic-based post-processing and hence \OurConvName cannot be used here as a drop-in replacement to further push its performance. In fact, STEm-Seg~\cite{Athar20ECCV} is the only method that uses 3D convolutions to incorporate temporal context, and as seen in Table.~\ref{tab:davis19-bench}, \OurConvName improves its performance from 63.4 to 64.6 $\mathcal{J} \& \mathcal{F}$.

\begin{table}[t]
\centering
\resizebox{\linewidth}{!}{%
\sisetup{detect-weight=true}
\begin{tabular}{l | c | cc }
\toprule
\multicolumn{4}{c}{DAVIS 2019 Unsupervised} \\
\midrule
Method & $\mathcal{J}\&\mathcal{F}$ Mean & $\mathcal{J}$ Mean& $\mathcal{F}$ Mean\\
\cmidrule(lr){1-4}
\texttt{\IRREL{$\text{KIS}^*$~\cite{Cho19CVPRW}}}  & \IRREL{59.9}&\IRREL{-} & \IRREL{-}  \\
\texttt{\IRREL{$\text{UnOVOST}^*$~\cite{Zulfikar19CVPRW}}}  &\IRREL{\textbf{67.0}} & \IRREL{\textbf{67.0}} & \IRREL{ \textbf{68.4} }\\
\texttt{\IRREL{RVOS}~\cite{Ventura19CVPR}} & \IRREL{41.2}& \IRREL{36.8} & \IRREL{45.7} \\
\texttt{\IRREL{AGNN}~\cite{Wang19ICCV}}  & \IRREL{61.1} & \IRREL{58.9}&\IRREL{63.2}  \\
\cmidrule(lr){1-4}
\texttt{STEm-Seg}~\cite{Athar20ECCV}    & 63.4 & 60.3 & 66.5 \\
STEm-Seg +\OurConvName  & \textbf{64.6} & \textbf{60.8} & \textbf{68.5} \\
\bottomrule
\end{tabular}
}
\caption{Results on the validation set of DAVIS'19 unsupervised VOS.}
\label{tab:davis19-bench}
\end{table}

\PAR{YouTube-VIS:}
We provide an overview of current methods for video instance segmentation on YoutubeVIS\cite{Yang19ICCV} in \reftab{sota_ytvis}.
Again, methods in gray do not use 3D convolutions.
The best performing method, MaskProp~\cite{Bertasius20CVPR}, achieves an impressive score of 46.6 mAP.
It extends Mask R-CNN~\cite{He17ICCV} with a mask propagation branch branch; there are no 3D convolutions which we can replace with \OurConvName in order to boost performance further.
STEm-Seg~\cite{Athar20ECCV} is the only method relying on 3D convolutions.
Replacing regular convolutions with \OurConvName in the decoder increases performance from 30.6 mAP to 32.3 mAP.
Despite a weaker ResNet50 backbone, STEm-Seg + \OurConvName is still competitive to many current architectures.

\begin{table}[t]
    \centering
    \begin{adjustbox}{max width=\textwidth}
    \begin{tabular}{l|c|cccc}
        \toprule
        Method & mAP & AP50 & AP75 & AR1 & AR10 \\
        \midrule
        \IRREL{FEELVOS}\cite{Voigtlaender19CVPR_FEELVOS} & \IRREL{26.9} & \IRREL{42.0} & \IRREL{29.7} & \IRREL{29.9} & \IRREL{33.4} \\
        \IRREL{IoUTracker+}~\cite{Yang19ICCV}     & \IRREL{23.6} & \IRREL{39.2} & \IRREL{25.5} & \IRREL{26.2} & \IRREL{30.9} \\
        \IRREL{OSMN}~\cite{Yang18CVPR}            & \IRREL{27.5} & \IRREL{45.1} & \IRREL{29.1} & \IRREL{28.6} & \IRREL{33.1} \\
        \IRREL{DeppSORT}~\cite{Wojke17ICIP}       & \IRREL{26.1} & \IRREL{42.9} & \IRREL{26.1} & \IRREL{27.8} & \IRREL{31.3} \\
        \IRREL{MaskTrack R-CNN}~\cite{Yang19ICCV} & \IRREL{30.3} & \IRREL{51.1} & \IRREL{32.6} & \IRREL{31.0} & \IRREL{35.5} \\
        \IRREL{SeqTracker}~\cite{Yang19ICCV}      & \IRREL{27.5} & \IRREL{45.7} & \IRREL{28.7} & \IRREL{29.7} & \IRREL{32.5} \\
        \IRREL{SipMask}~\cite{Cao20ECCV}          & \IRREL{32.5} & \IRREL{53.0} & \IRREL{33.3} & \IRREL{33.5} & \IRREL{38.9} \\
        \IRREL{CSipMask}~\cite{Qi21Arxiv}         & \IRREL{35.1} & \IRREL{55.6} & \IRREL{38.1} & \IRREL{35.8} & \IRREL{41.7} \\
        \IRREL{CMaskTrack R-CNN}~\cite{Qi21Arxiv} & \IRREL{32.1} & \IRREL{52.8} & \IRREL{34.9} & \IRREL{33.2} & \IRREL{37.9} \\
        \IRREL{CompFeat}~\cite{Fu21AAAI}          & \IRREL{35.3} & \IRREL{56.0} & \IRREL{38.6} & \IRREL{33.1} & \IRREL{40.3} \\
        \IRREL{VisTR (Res50)}~\cite{Wang21CVPR}   & \IRREL{36.2} & \IRREL{59.8} & \IRREL{36.9} & \IRREL{37.2} & \IRREL{42.4} \\
        \IRREL{VisTR (Res101)}~\cite{Wang21CVPR} & \IRREL{40.1} & \IRREL{\textbf{64.0}} & \IRREL{45.0} & \IRREL{38.3} & \IRREL{44.9} \\
        \IRREL{MaskProp}~\cite{Bertasius20CVPR} & \IRREL{\textbf{46.6}} & \IRREL{--} & \IRREL{\textbf{51.2}} & \IRREL{\textbf{44.0}} & \IRREL{\textbf{52.6}} \\
        \midrule
        STEm-Seg~\cite{Athar20ECCV}  & 30.6 & 50.7 & 33.5 & 31.6 & 37.1 \\
        STEm-Seg + \OurConvName  & \textbf{32.3} & \textbf{51.3} & \textbf{34.7} & \textbf{32.2} & \textbf{38.1} \\
        \bottomrule
    \end{tabular}
    \end{adjustbox}
    \caption{Performance comparison on the validation set of YoutubeVIS 2019~\cite{Yang19ICCV}. Baseline is STEm-Seg~\cite{Athar20ECCV} with a ResNet50 backbone.}
    \label{tab:sota_ytvis}
\end{table}

\begin{table}[t]
    \centering
    \begin{adjustbox}{max width=\textwidth}
    \begin{tabular}{l|c|cccc}
        \toprule
        Method & mAP & AP50 & AP75 & AR1 & AR10 \\
        \midrule
        \IRREL{CSipMask}~\cite{Qi21Arxiv}         & \IRREL{14.3} & \IRREL{29.9} & \IRREL{12.5} & \IRREL{\textbf{9.6}} & \IRREL{19.3} \\
        \IRREL{CMaskTrack R-CNN}~\cite{Qi21Arxiv} & \IRREL{15.4} & \IRREL{33.9} & \IRREL{13.1} & \IRREL{9.3} & \IRREL{20.0} \\
        \IRREL{CrossVIS~\cite{Yang21Arxiv}} & \IRREL{\textbf{18.1}} & \IRREL{\textbf{35.5}} & \IRREL{\textbf{16.9}} & \IRREL{--} & \IRREL{\textbf{--}} \\
        \midrule
        STEm-Seg~\cite{Athar20ECCV} & 14.3 & 31.5 & 12.4 & 10.2 & 20.7 \\
        STEm-Seg + \OurConvName  & \textbf{15.2} & \textbf{33.8} & \textbf{13.7} & \textbf{10.6} & \textbf{22.2} \\
        \bottomrule
    \end{tabular}
    \end{adjustbox}
    \caption{Performance comparison on the validation set of OVIS~\cite{Qi21Arxiv}.}
    \label{tab:sota_ovis}
\end{table}

\PAR{KITTI-MOTS:}
Recently, HOTA~\cite{Luiten21IJCV} has been proposed as a metric for tracking and segmentation.
We provide HOTA scores for our models in \reftab{kitti_mots_hota}, and compare our performance with Track R-CNN~\cite{Voigtlaender19CVPR}.
Our STEm-Seg baseline performs overall better than Track R-CNN; Track R-CNN provides a better detection accuracy (DetA in \reftab{kitti_mots_hota}), while STEm-Seg achieves a better association accuracy.
Both methods perform comparable in terms of localization accuracy.

\begin{table*}[t]
    \centering
    \begin{tabular}{l|cccc|cccc}
        \toprule
               & \multicolumn{4}{c|}{Car}      & \multicolumn{4}{c}{Pedestrian} \\
        Method & HOTA & DetA & AssA & LocA & HOTA & DetA & AssA & LocA \\
        \midrule
        Track R-CNN~\cite{Voigtlaender19CVPR} & 72.3 & \textbf{77.4} & 67.8 & 88.3 & 42.1 & \textbf{54.9} & 32.7 & 78.6 \\
        \midrule
        STEm-Seg~\cite{Athar20ECCV} & 73.1 & 68.6 & \textbf{78.2} & 88.7 & 47.9 & 48.8 & 47.2 & 79.6 \\
        STEm-Seg + DCNv1        & 73.3 & 70.4 & 76.7 & 88.8 & 45.5 & 46.6 & 44.8 & 78.5 \\
        STEm-Seg + DCNv2        & 72.7 & 70.0 & 75.9 & 88.7 & 47.7 & 47.8 & 48.1 & 78.9 \\
        STEm-Seg + \OurConvName & \textbf{74.1} & 70.5 & \textbf{78.2} & \textbf{89.4} & \textbf{50.1} & 50.3 & \textbf{50.3} & \textbf{80.0} \\
        \bottomrule
    \end{tabular}
    \caption{HOTA score on the validation set of KITTI MOTS. Baseline is STEm-Seg~\cite{Athar20ECCV} with a ResNet50 backbone.}
    \label{tab:kitti_mots_hota}
\end{table*}

\end{document}